\documentclass[10pt,twocolumn,letterpaper]{article}

\usepackage{cvpr}
\usepackage{times}
\usepackage{epsfig}
\usepackage{graphicx}
\usepackage{amsmath}
\usepackage{amssymb}
\usepackage{multirow}
\usepackage{subfigure}
\usepackage{textcomp}
\usepackage{float}
\usepackage{caption}
\usepackage{amssymb}
\usepackage{amsfonts}
\usepackage{multirow}
\usepackage{xcolor,colortbl}
\usepackage[pagebackref=true,breaklinks=true,letterpaper=true,colorlinks,bookmarks=false]{hyperref}

\def\cvprPaperID{****} 

\cvprfinalcopy 

\begin{document}

\title{Interpretable BoW Networks for Adversarial Example Detection}

\author{Krishna Kanth Nakka, Mathieu Salzmann\\
	CVLab, EPFL
}

\maketitle

\newif\ifdraft
\draftfalse

\definecolor{orange}{rgb}{1,0.5,0}

\ifdraft
\newcommand{\comment}[1]{}
\newcommand{\ms}[1]{{\color{blue}{#1}}}
\newcommand{\MS}[1]{{\color{blue}{\bf #1}}}
\newcommand{\kn}[1]{{\color{red}{#1}}}
\newcommand{\KN}[1]{{\color{red}{\bf #1}}}

\else
\newcommand{\comment}[1]{}
\newcommand{\MS}[1]{}
\newcommand{\ms}[1]{ #1 }
\newcommand{\KN}[1]{}
\newcommand{\kn}[1]{ #1 }
\fi

\newcommand{\bh}{\mathbf{h}}
\newcommand{\bz}{\mathbf{z}}
\newcommand{\bB}{\mathbf{B}}
\newcommand{\bV}{\mathbf{V}}
\newcommand{\bb}{\mathbf{b}}
\newcommand{\bX}{\mathbf{X}}
\newcommand{\bx}{\mathbf{x}}
\newcommand{\bbx}{\bar{\mathbf{x}}}
\newcommand{\by}{\mathbf{y}}
\newcommand{\bv}{\mathbf{v}}
\newcommand{\bI}{\mathbf{I}}
\newcommand{\bW}{\mathbf{W}}
\newcommand{\bw}{\mathbf{w}}
\newcommand{\bH}{\mathbf{H}}
\newcommand{\bs}{\mathbf{s}}
\newcommand{\bl}{\mathbf{l}}
\newcommand{\bpsi}{\mathbf{\psi}}
\newcommand{\Ekk}{\mathbb{E}}
\newcommand{\probkk}{\mathbb{P}}

\newcommand{\argmin}{\operatornamewithlimits{argmin}}
\newcommand{\argmax}{\operatornamewithlimits{argmax}}


\begin{abstract}
The standard approach to providing interpretability to deep convolutional neural networks (CNNs) consists of visualizing either their feature maps, or the image regions that contribute the most to the prediction. In this paper, we introduce an alternative strategy to interpret the results of a CNN. To this end, we leverage a Bag of visual Word representation within the network and associate a visual and semantic meaning to the corresponding codebook elements via the use of a generative adversarial network. The reason behind the prediction for a new sample can then be interpreted by looking at the visual representation of the most highly activated codeword. We then propose to exploit our interpretable BoW networks for adversarial example detection. To this end, we build upon the intuition that, while adversarial samples look very similar to real images, to produce incorrect predictions, they should activate codewords with a significantly different visual representation. We therefore cast the adversarial example detection problem as that of comparing the input image with the most highly activated visual codeword. As evidenced by our experiments, this allows us to outperform the state-of-the-art adversarial example detection methods on standard benchmarks, independently of the attack strategy.

	
\end{abstract}


\section{Introduction}

While the discriminative power of deep convolutional neural networks (CNNs) is nowadays virtually uncontested, one of the key research challenges that remain unaddressed is their interpretability; in many practical scenarios, one is not only interested in obtaining a prediction from the network, but also in understanding the reason behind this prediction. The main trends to tackle interpretability consist of post-training analysis to visualize either feature maps at different layers~\cite{visualizing1,visualizing2} or the image regions that contribute most to the decision~\cite{attpool1,gradcam}. 
In this paper, we introduce an alternative strategy to encode network interpretability via the use of Bag of visual Words (BoW) representations.

\begin{figure}[t]
	
	\centering
	\hspace{-0.2cm}\includegraphics[width=1.05\linewidth]{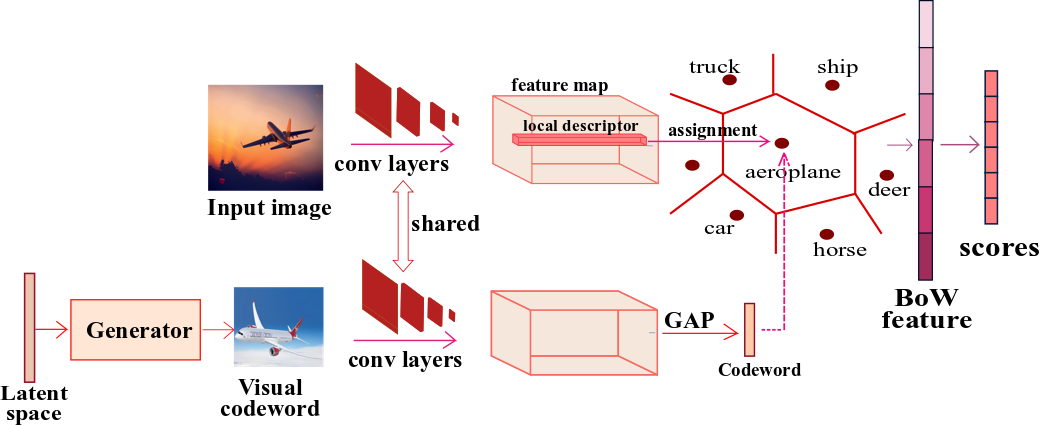}
	\vspace{-0.2cm}
	\caption{{\bf Interpretable BoW networks.} We use a GAN to generate a visual codebook, from which we obtain the codewords of a BoW model. This allows us to associate a visual and semantic meaning to the codewords, and, as shown in our experiments, to detect adversarial samples by comparing the input image with the most highly activated visual codeword.}
	\label{fig:teaser}
	\vspace{-8pt}
	
\end{figure}

BoW-based representations, such as histograms~\cite{bow1,bow2,bow3}, VLAD~\cite{vlad2012,allaboutvlad} and Fisher Vectors~\cite{fisher1, fisher2}, have a longstanding history in the computer vision community. In the current deep learning era, they have been re-visited to yield novel, structured ways of pooling local activations into a global image representation~\cite{netvlad2016,fishernet}. In essence, these strategies rely on measuring the distance between feature vectors and a codebook that is learnt jointly with the network parameters. Here, we argue that this distance-based representation is inherently interpretable; it allows one to understand the similarity between the input image and the codewords, which can be thought of as prototypes. However, in existing architectures, the codewords do not have a human-interpretable meaning, thus preventing one to visualize the reason behind the network's prediction.

In this paper, we therefore introduce an approach to provide  an interpretable meaning to the codebook of a BoW network. Specifically, we propose to associate a visual and semantic representation to each codeword. As illustrated in Fig.~\ref{fig:teaser}, this is achieved by making use of a 
generative adversarial network (GAN)~\cite{gan2014, dcgan2015} that maps latent vectors to images.
These images are then passed through the same base network as the one at the core of our BoW model, and we take the resulting feature vectors as our codebook elements. We then fine-tune the BoW model for classification with this codebook, which, in essence, allows it to learn a similarity between an input image and the images associated with the codewords. Ultimately, given a new image, observing the most highly activated codewords yields a visual and semantic interpretation of the network's prediction.

As a second contribution, we propose to leverage our interpretable BoW networks to detect adversarial examples, that is, images to which a small amount of structured noise has been added so as to fool a pre-trained network. To this end, we rely on the intuition that an adversarial example (i)~looks similar to the unaltered, but unknown, image; and (ii)~to produce an erroneous prediction, will have a high activation for a codeword that corresponds to the wrong class. The image associated with this codeword, however, constitutes a prototype of this wrong class, and thus will look very different from the adversarial example itself. We therefore cast adversarial example detection as the problem of comparing the adversarial image with the visual representation of the most highly activated codeword.

Our experiments demonstrate that our interpretable BoW networks (i) yield visually and semantically meaningful representations; (ii) are more robust than traditional CNNs to adversarial examples; and (iii) yield codewords whose visual representation can be effectively exploited to detect adversarial examples and even out-of-distribution examples. In particular, our approach outperforms the state-of-the-art adversarial example detection methods on standard benchmarks for state-of-the-art attack strategies, such as CW~\cite{CWattack}, FGSM~\cite{fgsm}, \comment{DeepFool~\cite{DF}} and BIM~\cite{bima}. We will make our code publicly available upon acceptance of the paper.

\comment{

Convolutional neural networks (CNNs) have achieved superior performance in many visual tasks, such as object classification and detection. Besides the discrimination power, model interpretability and uncertainty is another crucial property for neural networks. However, the interpretability of CNN's has presented challenges for years and typically studied in a post-training stage. In this paper, we focus on a new problem, i.e. without any additional human supervision, can we modify a CNN to make its top layers obtain interpretable knowledge representations? We expect the CNN to have a certain introspection of its representations during the end-to-end learning process so that the CNN can regularize its representations to ensure high interpretability.

We revisit the deep-structured architectures such as NetBoW and NetVLAD.  The structured layers in these architectures by design indirectly helps us to interpret the model decisions.  At the core of this layers is a codebook that is jointly learned with deep features in the training process. The distance-based feature representation in the structured layers can be used as a tool for model interpretability. However, in the present architectures, the codebook filters do not have human-level interpretations and not totally meaningful to understand the logic in decisions of CNNs.

Therefore, in this study, we aim to train the codewords that maintain a semantic attribute in the structured layer. We leverage a pretrained GAN to jointly learn the optimal visual dictionary for the task.  In this way, any image to the deep network is explained by its nearest codeword and more directly with the associated visual interpretation of the codeword.

The goal of this study can be summarized as follows.

\begin{itemize}
	
	\vspace{-5pt}
	\item We introduce interpretable semantic codewords into the deep-structured architectures by leveraging the expressiveness of  GAN to model the optimal visual dictionary.
	\vspace{-5pt}
	\item We analyze the reasons for misclassification in the settings of adversarial attacks and out-of-distribution (OOD) samples through activated codewords	
	\vspace{-5pt}
	
\end{itemize}


\textbf{Network interpretability:} The clear semantics in top layers of the network is of great importance when we need human beings to trust a network's prediction. In spite of the high accuracy of neural networks, human beings usually cannot fully trust a network, unless it can explain its logic for decisions, \emph{i.e.} what patterns are memorized for prediction. Given an image, current studies for network diagnosis localize image regions that contribute most to network predictions at the pixel level. Contrary to existing approaches, our work helps to understand the prediction logic of the CNNs through an interpretable structured representation in the top layers of the network.


\textbf{Contributions:} In this paper, we focus on end-to-end learning a CNN whose representations in structured layers, at the top of the network, are interpretable.  we propose a simple yet effective method to enforce codebook filters of structured layers to have semantic interpretability by generating them from real images.  Our high-level idea is the distance-based feature representation in the structured layers is by design a natural tool to study the interpretability of CNNs.  In order to enforce the codebook filters to be interpretable with a semantic class level attribute, we use of an unsupervised GAN that freely generate images.  The network is designed to learn the optimal  dictionary images from the large space of images generated by a GAN. The framework learns network weights together with an optimal visual dictionary that eventually leads to an interpretable structured representation. Experiments show that our approach has significantly improved the interpretability of CNNs in understanding the logic for its decisions to predict the label. We further explicitly compute the semantic dictionaries at every level of the network to analyze the reasons for prediction and study on predictive uncertainty.


Our framework is shown in Figure~\ref{fig:teaser} with the base network shared with input data and visual dictionary generated from the GAN. The input deep features of the BoW layer and its associated codebook filters are jointly learned. It is to be noted, we only optimize the latent space of the generator along with weights of the base network without retraining weights of a generator.  Once trained, we can remove the additional generator module to infer at the same cost as the standard structured networks.


<<<<<<< .mine
Recently, several works have shown that neural networks are known to be vulnerable to adversarial examples that can mislead to give incorrect predictions with high confidence. In this work, we focus to study the codebook activations of adversarial examples in comparison to their clean counterparts. Our experiments conclude that real images activate codewords that are semantically close to its real label while the adversarial images activate a different codeword that explains the adversarial label. We leverage this information to build an adversarial detector to detect malicious samples in case of adversarial settings. We also found that our proposed framework with semantic dictionaries can be used to detect the OOD samples without explicitly tuning with them.  We further show that our adversarial detector trained on FGSM attack can generalize well to other attacks.||||||| .r171
Recently, several works have shown that neural networks are known to be vulnerable to adversarial examples that can mislead to give incorrect predictions with high confidence. In this work, we focus to study the codebook activations of adversarial examples in comparison to their clean counterparts. Our experiments conclude that real images activate codewords that are semantically close to its real label while the adversarial images activate a different codeword that explains the adversarial label. We leverage this information to build an adversarial detector to detect malicious samples in case of adversarial settings. We also found that our proposed framework with semantic dictionaries can be used to detect the OOD samples without explicitly tuning with them.  We show that our adversarial detector tuned with only using in-distribution (training) samples while maintaining its performance. We further show that our adversarial detector trained on FGSM attack can generalize well to other attacks.=======
Recently, several works have shown that neural networks are known to be vulnerable to adversarial examples that can mislead to give incorrect predictions with high confidence. In this work, we focus to study the codebook activations of adversarial examples in comparison to their clean counterparts. Our experiments conclude that real images activate codewords that are semantically close to its real label while the adversarial images activate a different codeword that explains the adversarial label. We leverage this information to build an adversarial detector to detect malicious samples in case of adversarial settings. We also found that our proposed framework with semantic dictionaries can be used to detect the OOD samples without explicitly tuning with them.  We show that our adversarial detector tuned with only using in-distribution (training) samples while maintaining its performance. We further show that our adversarial detector trained on FGSM attack can generalize well to other attacks.

}



\section{Related Work}

\noindent\textbf{BoW representations.} Automatically recognizing 
objects in images and videos is central to a wide variety of application domains, such as security and autonomous driving. Visual recognition has therefore been one of the fundamental goals of computer vision since its inception. Prior to 2012, most methods followed a two-step pipeline comprised of handcrafted feature extraction from the images and separately training a classifier. Bags of visual Words (BoW)~\cite{bow1,bow2,bow3}, that is, histograms extracted by comparing local features to the elements of a codebook obtained from the training data, quickly became popular handcrafted representations. 
They were then extended to VLAD~\cite{vlad2012} and Fisher Vectors~\cite{fisher1, fisher2}, which encode higher-order statistics of the data with respect to the codewords. After  AlexNet's impressive performance~\cite{alexnet},  much of the visual recognition research moved to employing deep networks. While most architectures extract the final image representation using standard convolutions, a few works attempted to leverage  knowledge from the handcrafted features era. In particular,~\cite{multiscale} exploits a VLAD pooling strategy on features extracted with a pre-trained network. While this separates feature extraction and classifier training, HistNet~\cite{histnet2018}, NetVLAD~\cite{netvlad2016}, ActionVLAD~\cite{actionvlad}, and Deep FisherNet~\cite{fishernet} constitute end-to-end learning frameworks leveraging BoW, VLAD and Fisher Vector representations, respectively. Here, we contribute to this effort with a new deep architecture that incorporates the notion of interpretability in the BoW representation.

\noindent\textbf{Interpreting CNNs.}
Attempts at interpreting the representations learned by CNNs or their predictions remain few, and existing methods follow two main trends. The first one~\cite{visualizing1,visualizing2,visualizing3,visualizing4} focuses on visualizing the CNN filters in a post-training stage, by either inverting the network, or performing gradient ascent in image space to maximize neuron's activations. The second trend consists of identifying the image regions that bear the most responsibility for the prediction. Note that this idea can be traced back to non-deep learning strategies, such as image representations based on object detectors~\cite{objectbanks} and classemes~\cite{classesmes}, and even part-based models~\cite{felz_object2008,felz_object2010,felz_pictorial}. In the deep learning context, this was introduced by~\cite{CAM}, extended in~\cite{gradcam}, both of which use a post-training strategy. This was followed by~\cite{residualattention} that incorporates an attention module at every layer of the network. In~\cite{interpretableCNN}, an additional loss function was designed to assign each CNN filter to an object part. Recently~\cite{attpool1,nakka2018} have proposed to leverage attention maps during pooling operations. In this paper, we introduce an alternative way to interpret the prediction of a CNN by providing a visual and semantic representation to the codewords of a BoW model. While~\cite{nakka2018} also relies on a histogram-based representation, their approach differs fundamentally from ours, in that it does not assign a visual interpretation to the codebook.


\noindent\textbf{Adversarial attacks and detection.}
When the sensitivity of deep networks to adversarial attacks was identified~\cite{adv2013}, initial works focused on developing defense strategies~\cite{fgsm,bima}, aiming to robustify the networks.
However, these defenses were typically found to be vulnerable to optimization-based techniques~\cite{CWattack}. Therefore, the research focus has increasingly shifted towards detecting adversarial samples, thus allowing one to discard them instead of attempting to be robust to them. In this context,~\cite{metzen2017} proposed to use a separate subnetwork to detect adversarial examples;~\cite{grosse2017} relies on knowledge distillation and Bayesian uncertainty to train a simple logistic regression detector;~\cite{ma2018characterizing} exploits a measure of local intrinsic dimensionality to identify the adversarial examples. Here, we show that we can outperform all these methods by learning to compare the input image with the visual representation of the most highly activated codeword in our interpretable BoW network.


Another problem related to adversarial sample detection is that of identifying out-of-distribution (OOD) examples. This task has been addressed by training a detector on the softmax scores of a network~\cite{OODbaseline}, extended in~\cite{ODIN} by an additional pre-processing of the network input. The contemporary work~\cite{Mahalanobis2018} introduced a unified framework for adversarial and OOD sample detection based on the Mahalanobis distance between hidden features and their class-conditional distributions. Here, unlike existing methods, we base our adversarial and OOD detection framework directly on a visual and semantic interpretation of the network's prediction. This, we believe, would further give one the possibility to visually analyze the successful attacks, opening the door to human intervention in the detection process, while facilitating the human's task.



\begin{figure*}[htp]
	
	\centering
	\includegraphics[width=\linewidth, height=5cm]{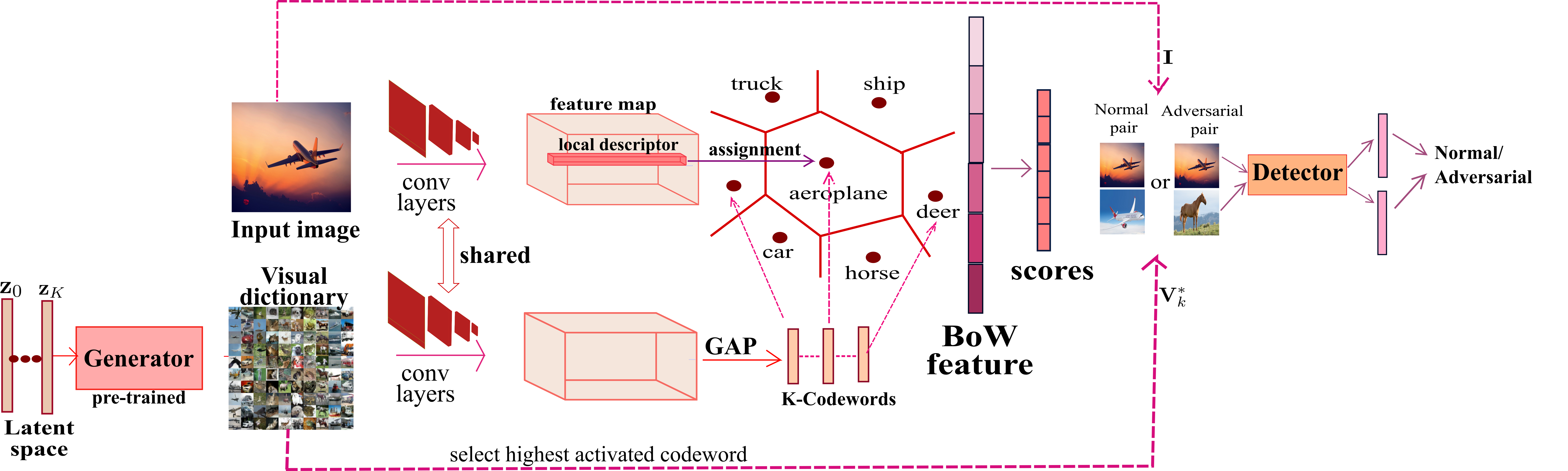}
	\caption{{\bf Interpretable BoW network for adversarial sample detection.} An input image is passed through the network, and we retrieve the visual codeword with highest activation. We then pass these two images through a two-stream model which compares them using the Euclidean distance in feature space, so as to determine if they belong to the same class.}
	\label{fig:algorithm}
	
\end{figure*}

\section{Method}

In this section, we first introduce our interpretable BoW networks.  We then show how to use the resulting visual codebook as a cue to detect adversarial and out-of-distribution (OOD) samples.

\subsection{Interpretable BoW Networks}
\label{sec:IBoW}
Typically, deep neural networks aggregate the local features of the convolutional maps, including the last one, using average or max pooling, which does not separately account for the contributions of the individual image regions. By contrast, here, we aim to leverage a BoW representation, where these individual contributions are preserved by comparing each local feature to the elements of a codebook. More specifically, we draw our inspiration from the histogram-based models of~\cite{histnet2018,netvlad2016,fishernet}, but propose to provide an interpretable representation of the codebook. Below, we first formalize the BoW network, and then introduce our interpretable representation.



\vspace{0.1cm}
\noindent\textbf{BoW network.} Formally, let $\bI$ be an image input to a CNN, and $\bX \in \mathbb{R}^{W \times H \times D}$ be the feature map output by the CNN's last convolutional layer, with spatial resolution $W \times H$ and $D$ channels. $\bX$ can be thought of as $M = W \cdot H$ local descriptors $\bx_{i}$ of dimension $D$. We then introduce a BoW layer that, given a codebook $\bB$ with $K$ codewords, produces a $K$-dimensional representation of the form
\begin{equation}
\bh(\bX) = \frac{1}{M} \sum_{i=1}^M \bh_i(\bx_i)\;,
\label{eq:bow_vector}
\end{equation}
which aggregates local histogram-like vectors for each feature $\bx_i$. We express these local vectors as
\begin{equation}
\bh_i(\bx_i) = [a_0(\bx_i), a_1(\bx_i), \cdots, a_K(\bx_i)]^T\;,
\label{eq:vlad_vector}
\end{equation}
where $a_k(\bx_i)$ is given by
\begin{equation}
a_k(\bx_{i})= \frac{ e^{-\alpha\left\| \bx_{i}-\bb_k  \right\|^2}}{\sum_{k'}{e^{-\alpha\left\| \bx_{i}-\bb_{k'} \right \|^2}}}\;.
\label{eq:vlad_soft_assgn}
\end{equation}    
This value represents the assignment of descriptor $\bx_{i}$ to codeword $\bb_k$. Note that, in the classical BoW formalism, the assignments are binary, with each descriptor being assigned to a single codeword. Within a deep learning context, for differentiability, we relax them as soft assignments, with $\alpha$ a hyper-parameter defining the softness. The resulting BoW vector $\bh(\bX)$ then acts as input to the final classification layer of the network. 

To obtain the codebook $\bB$, one typically first trains the base network, with global average pooling (GAP) of $\bX$ and a softmax classifier, and then performs $K$-means clustering on the local features $\{\bx_i\}_{i=1}^M$ of all training samples~\cite{netvlad2016,fishernet}. Since the codebook is then defined over abstract features, it does not have a clear visual or semantic meaning. Here, we propose to provide the codebook with such a meaning, which then translates to making our BoW network interpretable; given an input image $\bI$, one can analyze the prediction by visualizing the representation of the most highly activated codeword.


\vspace{0.1cm}
\noindent\textbf{Providing interpretability.} To obtain a visual and semantic codebook representation, we propose to exploit the generator of a pre-trained GAN. As illustrated in Fig.~\ref{fig:teaser}, the images obtained from this generator are passed through the same base network as that of our BoW model, and the resulting average-pooled features taken as codewords. As a result, the set of $K$ generated images $\{\bV_k\}_{k=1}^K$ can be thought of as a visual dictionary, and each codeword $\bb_k$ in the codebook $\bB$ is directly associated to a visual interpretation $\bV_k$. As GANs are now able to generate diverse images that cover all the classes in a dataset, this visual interpretation inherently comes with a semantic meaning. 

Given an input image $\bI$, with corresponding final feature map $\bX$, we can obtain an interpretation of its prediction by observing the visual codeword $\bV_{k^*}$ associated to the most highly activated codeword, that is, 
\begin{equation}
k^* = \argmax_{k}\bh^k(\bX)\;,
\label{eq:max_act}
\end{equation}
where $\bh^k(\bX)$ denotes the $k^{th}$ element of the vector $\bh$ in Eq.~\ref{eq:bow_vector}. For a sample from class $c$, this visual codeword will typically correspond to an image of the same class. It will, moreover, depict characteristics similar to that of the input image, and, as shown in our experiments, different codewords from the same class $c$ will focus on different characteristics, thus truly acting as prototypes.

\vspace{0.1cm}
\noindent\textbf{Training.}
In principle, our model can be trained in an end-to-end manner. We found, however, that training all the different parts jointly from scratch was unstable. To overcome this, we rely on the following training strategy. We first pre-train the GAN on the training images of the dataset of interest, and train the base network of our BoW model using average pooling with a standard softmax classifier. We then compute an initial codebook $\bB_0$ via $K$-means clustering of the average-pooled features of the last convolutional layer
of this base network. Specifically, we define an equal number of codewords $S$ for each class. Thus, for a $C$-class dataset, we obtain $K=CS$ codewords, and force each group of $S$ codewords to come from the same class during the clustering procedure. Note that this further encodes the notion of semantic meaning of each codeword.

To obtain a visual representation of each codeword in $\bB_0$, we optimize the input $\bz$ of the generator, whose weights are fixed, so as to generate an image that, when passed through the base network, yields features that are close to the codeword in the least-square sense. That is, formally, for each codeword $\bb_{0,k}$ in $\bB_0$, we solve
\begin{equation}
\bz^*_k = \argmin_{\bz} \|f(g(\bz)) - \bb_{0,k}\|^2\;,
\end{equation}
where $g(\cdot)$ represents the generator, and $f(\cdot)$ the base network up to the last average pooling operation. We then take our codebook $\bB$ to be the set of generated features $\{f(g(\bz^*_k))\}$, and train our BoW model by replacing the base network classification layer with a layer that maps the BoW representation to the class labels and using the standard cross-entropy loss. 

Note that, while this training procedure may seem costly, this has no effect on the computational cost at test time. Indeed, inference only involves a forward pass through our BoW network, which, compared to the base network, only requires us to store the additional codebook $\bB$; that is, the generator network is not needed anymore.

\comment{
In this regard, we use a deep generative network ~$G$ to model the semantic codebook as shown in Figure~\ref{fig:algorithm}. The images obtained from the generator is passed through the base network and the resulting features are considered as codeword elements.  In this way, we can understand what each codeword $b_j$ by means of it visual representation  $v_j $.  In order to learn optimal and capture the high variance in the codewords, we use a generator trained with GAN objective that can freely synthesize images of the dataset from a latent distribution with a great expressive power.  The joint network converges to find its best approximation of codewords for the dataset at hand. At the end of the training phase, the base network together with generator produces codewords that are not only interpretable but also encodes the data set characteristics through a visual dictionary $V$. This modification to codewords of the BoW layer can be assumed as an additional semantic prior to codeword filters in addition to standard recognition objective.

By training codewords with additional semantic prior, we have the flexibility of interpreting its learned structured representation.  Typically, the neural network has a final fully connected classification layer whose values encodes the probability the image belongs to a  certain class. While the score can be interpreted as the confidence of assignment to that label. It falls short in explaining the logic that led to this decision. However, in our present setup with interpretable BoW layer, its representation can give us a sense of understanding of model decisions through its highest activated codewords and more directly with associated visual interpretation from the generator network

Our complete framework, as shown in Figure~\ref{fig:algorithm}.  When a sample input of class $c$ is passed through the network, it activates the codeword $\bb_j$ in $\bB$ that is semantically close to the input image.  In other words, the deep feature representation of image $\bI$  assigned to codeword $\bb_j$ corresponding to the image $\bv_j$ of visual dictionary $\bV$.  These codewords represent the anchors using which the model predicts the output probabilities using the classification layer.

Also, it has to be noted although the training of joint network (base classifier with the generator network ) is costly at the training stage, it has no effect during inference.  The codewords filters $\bb_j$ post training can be used directly to predict the output label without any access to the generator network.

Further, we can also construct semantic dictionaries at each layer of the network and understand its internal hidden representation through its activated semantic codewords.
}

\subsection{Detecting Adversarial Examples}
\label{sec:method_detection}
By providing a visual and semantic meaning related to the network's prediction, our interpretable BoW network can be leveraged to detect adversarial examples. The reasoning behind this is the following: Typically, adversarial attacks aim to add the smallest amount of perturbation to an image so that the network misclassifies it. While this perturbation should be imperceptible to humans, it should strongly affect the resulting deep representation. In our case, this representation is the BoW one, and, for misclassification to occur, the most highly activated codeword should typically be associated to the wrong class. As such, its visual representation should look significantly different from the adversarial example. We therefore propose to train an adversary detector network that, given two input images, predicts whether they belong to the same class or not.

Formally, our detection framework, depicted by Fig.~\ref{fig:algorithm}, proceeds as follows. An image $\bI$, adversarial or not, is passed through our interpretable BoW network, and we retrieve the visual codeword $\bV_{k^*}$ corresponding to the highest activation using Eq.~\ref{eq:max_act}. We then pass $\bI$ and $\bV_{k^*}$ to our adversary detector, which outputs a binary label indicating whether the two images belong to the same class or not. If they don't, then $\bI$ is detected as an adversarial example.

Our detector is a two-stream network that extracts features for the two images independently. To train this network, we make use of the contrastive loss~\cite{CLoss} , which aims to make  the Euclidean distance between pairs of mismatched images larger than a margin $m=1$, while minimizing that of matching pairs. Detection is then performed by comparing the Euclidean distance to the margin.

While, in our experiments, we train the detector to identify adversarial examples, it can also be used to detect OOD samples, even without any re-training. In essence, the intuition remains unchanged: An OOD sample will activate a codeword whose visual representation looks different from the input image. The detection procedure is thus the same as in the case of adversarial examples.

\comment{
While the above interpretable BoW network provides insights on network decisions, it has interesting applicability on understanding its decisions in the case of adversarial settings. Typically, the attacker targets the system to misclassify with least amount of perturbation that is imperceptible to humans in the input image space, but the resulting deep representation is completely different from its expected normal representation. We believe our proposed semantic dictionaries are a natural candidates to understand the network decisions in such adversarial settings and further build a detection system leveraging its interpretable BoW representation.
 
To this end, we study the deep representation of images in adversarial settings. For example, a normal image $\bI_r$ with a label $y_r$ is attacked to predict an erroneous output label $y_a$.  Consider that a normal image $\bI_r$  activates a codeword $\bb_r$ corresponding to visual element $\bv_a$ and its adversarial counterpart $\bI_a$ activates codeword $\bb_a$ associated with visual word  $\bv_a$. Although, $\bI_r$ and $\bI_a$ are visually similar, its associated visual codewords $\bv_r$ and $\bv_a$ are semantically dissimilar. 

We use this cue to build a  detector to find the inconsistencies between its visual representation of activated codeword and the input image to the network. The detector is branched out from the main classifier network to predict the similarity between the input the visual interpretation of highest activated codeword. We conjecture such a detection system can be a simple model as its task at hand to predict similarity is relatively easy than a standard task to train all possible pairs. This is because the positive and negative pairs are well separated inherently. The positive pairs from the normal samples are highly similar in image space while the negative pairs from the adversarial images on the contrary are strongly dissimilar making the detection easy.

\comment{
In order to fool the BoW network, the attacker has to alter its deep feature representation such that its assigned codeword belongs to completely different semantics. 
}

While the detector is specifically trained to learn the adversarial samples, it can be additionally used to detect a test-sample from out of distribution dataset.  As in the case of adversarial detection, we compare the input OOD image with highest activated visual codeword to predict the similarity score. We observed a detector validated with only adversarial and clean images of in-distribution can be used to detect out of distribution samples. While the earlier works in adversarial and OOD detection imposes certain preprocessing steps to generate separable features by adding noise to input image, our approach automatically learns such a representation by associating each codeword to a visual image thereby producing a more discriminative inputs to detection system.
}


\section{Experiments}

We now empirically evaluate our interpretable BoW networks. To this end, we first analyze the semantic meaning of the learned visual dictionaries. We then demonstrate their benefits to detect adversarial and out-of-distribution samples. In these experiments, we used the standard benchmark datasets employed for adversarial sample detection, that is, MNIST~\cite{mnist}, F-MNIST~\cite{fmnist}, CIFAR-10~\cite{cifar10}, SVHN~\cite{netzer2011reading}. 

\vspace{0.1cm}
\noindent\textbf{Implementation details.}  For our comparisons to be meaningful, we rely on the same base architecture as in~\cite{ma2018characterizing} on each dataset. This architecture is first trained with a softmax classifier for 50-100 epochs and used to compute the initial codebook $\bB_0$. In parallel, we train either a GAN~\cite{gan2014} or a WGAN~\cite{WGAN-GP}, depending on the dataset, for 100k iterations. From the resulting generator and $\bB_0$, we obtain our interpretable codebook $\bB$ following the procedure described in Section~\ref{sec:IBoW}. We then train the classification layer of our interpretable BoW model for $40$ epochs.

\comment{
For our comparisons to be meaningful, we rely on the same base architecture as in~\cite{ma2018characterizing} on each dataset. This architecture is first trained with a softmax classifier for 50-100 epochs. We then compute the initial codebook $\bB_0$ of our BoW network via $k$-means clustering of the features of the penultimate layer of this base network. Specifically, we define an equal number of codewords $k$ for each class. Thus, for a $C$-class dataset, we obtain $K=Ck$ clusters codewords. In all our experiments, we generate $k=50$ clusters per class.

For each dataset, we then train a GAN~\cite{gan204NIPS, WGAN-GP} \MS{Do you use the standard GAN of Goodfellow? If not, we should give the correct reference.}\KN{Both GAN and WGAN are used depending on dataset, GAN goodfellow reference corrected} for 100k iterations. To obtain a visual representation of each codeword in $\bB_0$, we optimize the input $\bz$ of the generator, whose weights are fixed, so as to generate an image that, when passed through the base network, yields features that are close to the codeword in the least-square sense\MS{Correct?} \KN{True}. Given this initialization, the fixed generator and the base network, we then train our BoW model by replacing the base network classification layer with a layer that maps to the BoW representation to the class labels. \MS{The only thing that is optimized is this layer, or do you also optimize the $\bz$ variables?} \KN{Only classification layer} To this end, we make use of the cross-entropy loss, and train our BoW model for $40$ epochs.
}	
	

In all our experiments, we set $\alpha= 100$ in the BoW soft-assignment policy of Eq.~\ref{eq:vlad_soft_assgn}, and use the Adam~\cite{adam} optimizer with a learning rate of 0.001 and a decay rate of 0.1 applied every 20 epochs. Due to space limitation, we provide the detail of the base classifier and the GAN architecture in the supplementary material. The test errors on MNIST, FMNIST,  CIFAR-10, and SVHN using the softmax classifier and the BoW one are given in Table~\ref{tbl:classification_acc}. In essence, both classifiers perform on par. Note that our goal here is not to advocate for superior performance of the BoW model, but rather for its use to provide a visual interpretation, as discussed below.

\begin{table}[t]
	\centering
	\begin{tabular}{|c|c|c|}
		\hline
		Dataset & Base network & BoW network  \\
		\hline
		MNIST      & 99.23 & 99.15   \\ 
		FMNIST   &         92.46  & 92.06  \\ 
		CIFAR-10  &      87.54     & 87.95  \\ 
		SVHN       &        92.43  & 91.81  \\ 
		\hline
	\end{tabular}
	\vspace{-0.1cm}
	\caption{{\bf Classification accuracy (in \%) of the base and BoW networks.} Note that both models perform on par, but the BoW one allows us to obtain a visual interpretation.}
	\label{tbl:classification_acc}
\end{table}

\subsection{Visualizing BoW Codewords} \label{sec:exp_semdict}

\begin{figure} [t] 
	\subfigure[MNIST codewords]
	{
		\begin{tabular}{@{}c@{}}
			\includegraphics[width=.46\linewidth]{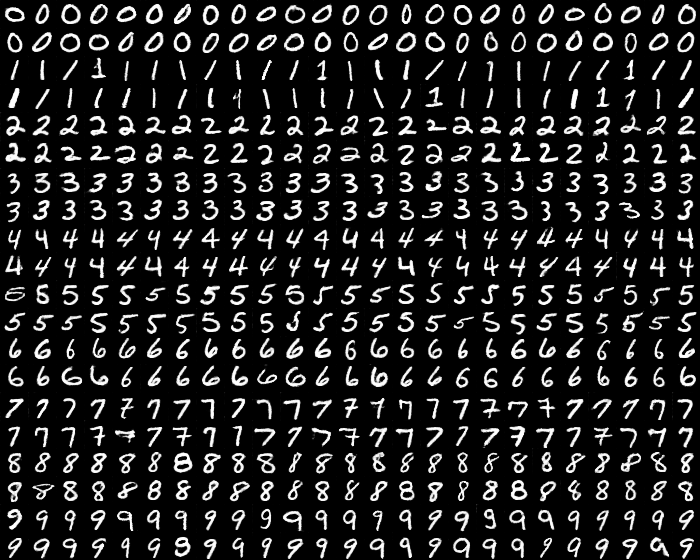} 
		\end{tabular}
		\label{codewords:a}
	} 
	\,
	\subfigure[FMNIST codewords]
	{
		\begin{tabular}{@{}c@{}}
			\includegraphics[width=.46\linewidth]{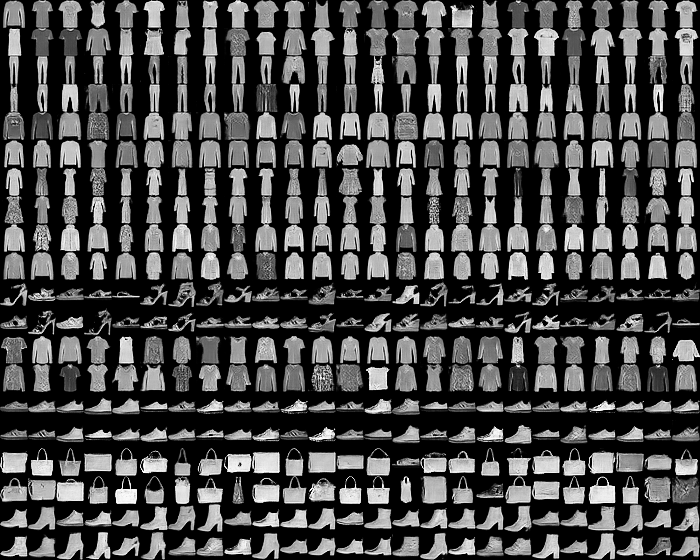} 
		\end{tabular}
		\label{codewords:b}
	} 
	\vspace{-0.1in}
	\subfigure[SVHN codewords]
	{
		\begin{tabular}{@{}cccc@{}}
			\includegraphics[width=.46\linewidth]{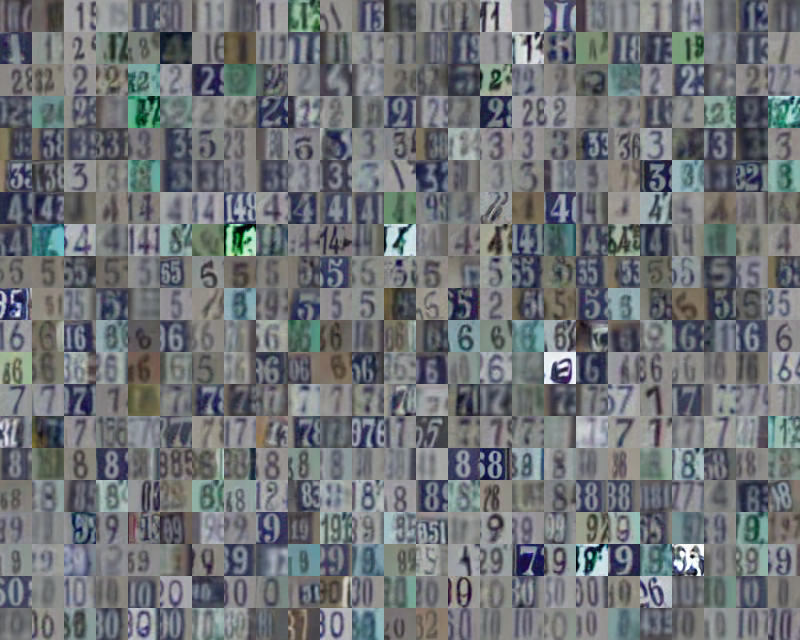} 
		\end{tabular}
		\label{codewords:c}
	}
	\,
	\subfigure[CIFAR-10 codewords]
	{
		\begin{tabular}{@{}cccc@{}}
			\includegraphics[width=.42\linewidth]{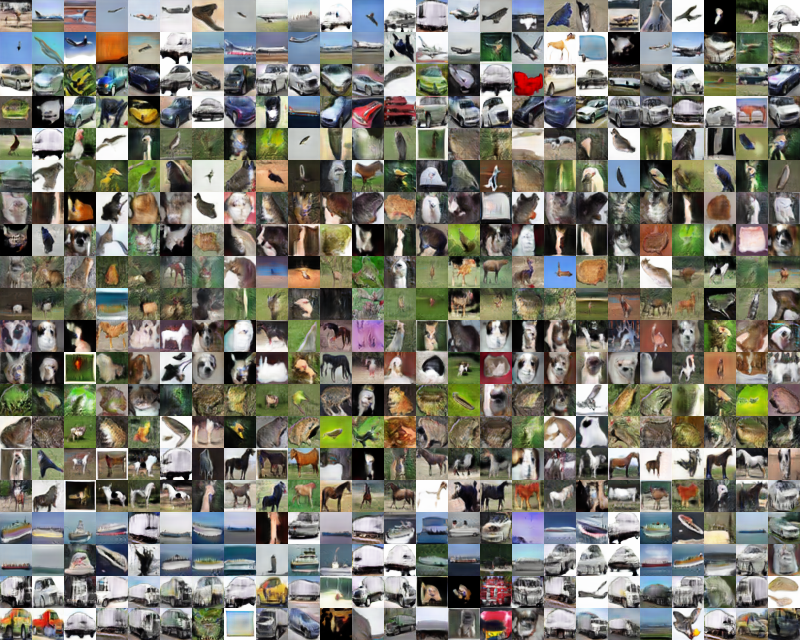}
		\end{tabular}
		\label{codewords:d}
	}
	\caption{{\bf Visual interpretations of the BoW codebooks.} Every two rows correspond to one class. Note that these codewords capture the diversity of the classes in each dataset (better seen by zooming in).}
	\label{codewords}
\end{figure}

Each codeword in our BoW model is directly associated with an image generated by a GAN.  
In Fig.~\ref{codewords}, we visualize these images for MNIST, FMNIST, SVHN and CIFAR-10.  Zooming in on the figure confirms that the learned codebooks nicely cover the diversity of the classes in these datasets. For example, in FMNIST, each garment appears in a variety of sizes and styles. Similarly, in CIFAR-10, each object appears in different colors, orientations and in front of different backgrounds. Furthermore, and more importantly, each one of these images retains the semantic meaning of the class label for which it was generated. This will prove key to the success of our adversarial sample detector, as discussed in the next section.


In the top row of Fig.~\ref{fig:highest_codewords}, we show the codewords with highest activation for a few correctly-classified images of each dataset. 
Note that, in most cases, the corresponding codeword has the same semantic meaning as the input image, not only because it corresponds to the same class, but also because it depicts a visually similar content, e.g., in terms of color and orientation. More such visualizations are provided in the supplementary material.


\comment{
\begin{figure} [t] 
	\subfigure[Activated  codeword for normal images]
	{
		\begin{tabular}{@{}c@{}}
			\includegraphics[width=.46\textwidth]{figs/mnist_codewords.png} 
		\end{tabular}
		\label{fig:highest_codewords_clean}
		
	} 
	\,
	\subfigure[Activated  codeword for adversarial images]
	{
		\begin{tabular}{@{}c@{}}
			\includegraphics[width=.46\textwidth]{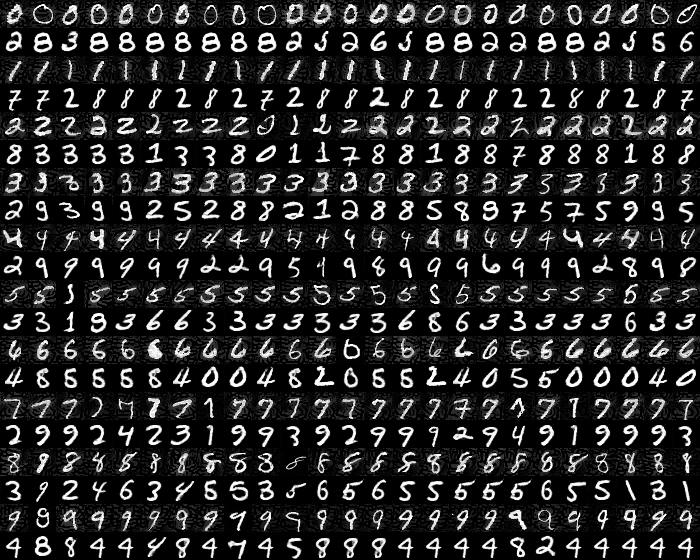} 
		\end{tabular}
		\label{fig:highest_codewords_adv}
	} 
	
	\caption{{\bf Most highly activated codewords for normal and adversarial samples (BIM-a attack).} In every pair of rows, we show the input image (top) and the corresponding visual codeword (bottom). Visualizations for other datasets and attacks are provided in the supplementary material. \MS{We may want to separate this into two figures?}}
	\label{fig:highest_codewords}
	
\end{figure}

}

\begin{figure*}[htp]
	
	\includegraphics[width=\linewidth, height=7cm]{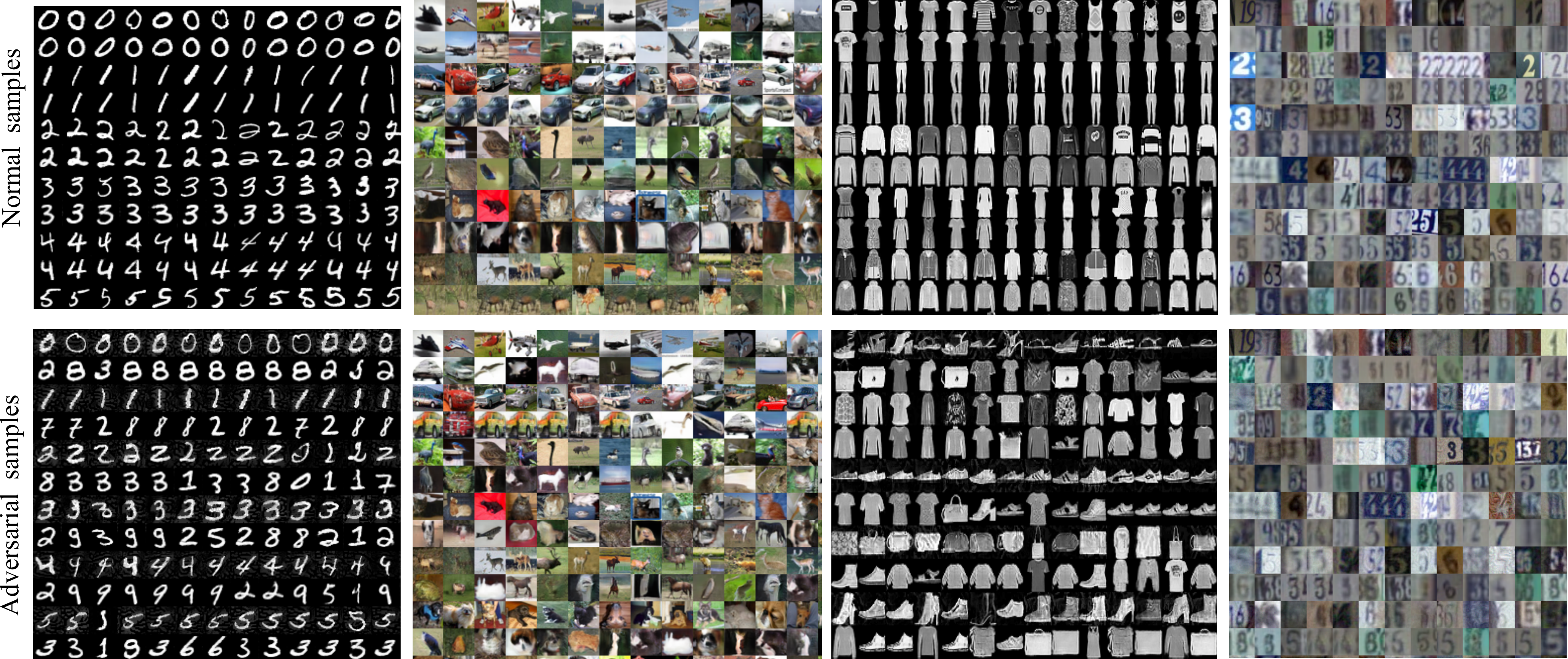}
	\caption{{\bf Visualization of the most highly activated codeword for normal and adversarial samples.} Each two rows within a block show the input image (top) and corresponding codeword (bottom). The adversarial examples were obtained with a BIM-a attack. Additional visualizations for other datasets and attacks are provided in the supplementary material. }
	\label{fig:highest_codewords}
	
\end{figure*}

\comment{

\begin{figure*}[htp]
	
	\centering
	\includegraphics[width=\linewidth, height=3cm]{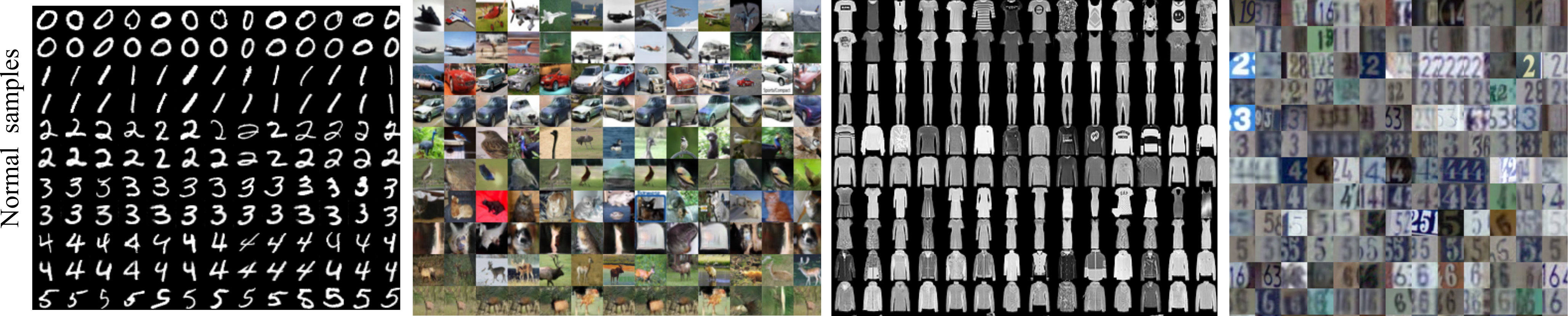}
	\caption{Visualization of highest activated codeword for normal and adversarial samples (BIM-a attack). \kn{Each image block has sequence of two rows with first row corresponding to input image and second row corresponds to activated codeword (bottom)}. Please refer appendix for more visualizations on other datasets and attacks. \MS{We may want to separate this into two figures?}}
	\label{fig:highest_codewords}
	
\end{figure*}

\begin{figure*}[htp]
	
	\centering
	\includegraphics[width=\linewidth, height=3cm]{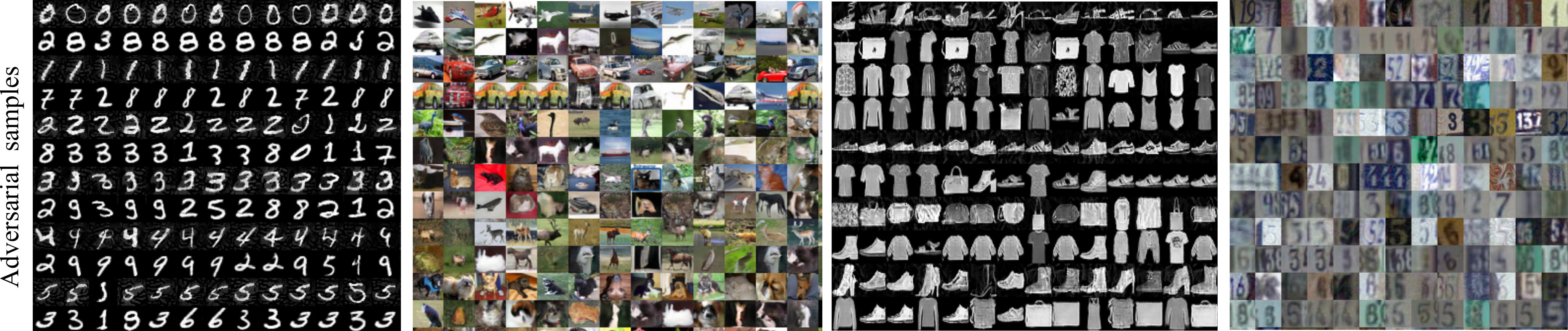}
	\caption{Visualization of highest activated codeword for adversarial samples (BIM-a attack). \kn{Each image block has sequence of two rows with first row corresponding to input image and second row corresponds to activated codeword.}  \MS{We may want to separate this into two figures?}}
	\label{fig:highest_codewords}
	
\end{figure*}

}

\subsection{Adversarial Attacks} \label{sec:exp_advdetect}

We make use of the state-of-the-art attack methods, FGSM~\cite{fgsm}, BIM-a~\cite{bima}, BIM-b~\cite{bima}, DeepFool~\cite{DF} and CW~\cite{CWattack}, to evaluate (i) the robustness of our interpretable BoW model to adversarial samples; and (ii) the effectiveness of our detection strategy. In both cases, following common practice~\cite{ma2018characterizing}, we discard the images that were misclassified by the original networks from this evaluation. 

To first validate our intuition that adversarial samples will activate codewords corresponding to the wrong classes, and that these codewords will be visually dissimilar to the input image, in the bottom row of Fig.~\ref{fig:highest_codewords}, we provide the most highly activated codeword images for a few successful adversarial attacks. Note that these images are indeed semantically and visually different, which will facilitate the task of our detector.



\subsubsection{Robustness of Interpretable BoW Networks} 
Before evaluating the effectiveness of our detection strategy, we study the robustness of our BoW model to the adversarial attacks. In particular, we focus on white-box attacks, where the attacker has access to the exact model it aims to fool. There are two ways to attack our BoW model: One can generate adversarial examples either for the BoW model itself, or for the base network. We refer to the latter as transferred BoW (T-BoW) attacks. In Table~\ref{table:asr}, we compare the success rates of different attack strategies on our BoW model with those on the base network. Note that attacks directly targeting our BoW model  are significantly less successful than those on the base network. While transferring the adversarial samples of the base model to our BoW network is more effective than the direct attacks, the success rates remain lower than on the base model. Note that, as shown in the supplementary material, the $L_2$ norm of the perturbations was similar for all network settings. This, we believe, shows that our BoW models are more robust than traditional networks to adversarial attacks. Nevertheless, some attacks are successful, and we now turn to the problem of detecting them.

\begin{table}[t!]

	\resizebox{\linewidth}{!}{%
		
		\begin{tabular}{|c|c|cccc|}
			\hline
			\multirow{1}{*}{Dataset}  & Model & FGM & BIM-a & BIM-b  & CW  \\ \hline
			
			\multirow{3}{*}{MNIST} 
			& Base    & 93.68 &100.0 &100.0  & 100.0 \\
			&  T- Bow  & 92.70 & 38.94 & 100.0 &  31.40 \\ 
			&   BoW &   \textbf{39.23} & \textbf{23.96} & \textbf{23.96} &   \textbf{3.9} \\ \hline
			
			\multirow{3}{*}{F-MNIST} 
			& Base    & 99.64 &100.0 &100.0  & 100.0 \\
			&  T- Bow  &  98.80 & 65.52 & 100.0  &\textbf{49.08} \\ 
			&   BoW &   \textbf{57.78} & \textbf{32.09} & \textbf{31.97} & 66.78 \\ \hline

			\multirow{3}{*}{SVHN} 
			& Base   & 97.0 & 100.0 & 100.0 &100.0  \\
			&  T- Bow & 96.83 &93.77 & 100.0 & \textbf{75.24}   \\ 
			&   BoW &  \textbf{73.39} & \textbf{73.11} & \textbf{72.24} & 86.42 \\ \hline
			
			\multirow{3}{*}{CIFAR-10} 
			& Base  &82.70 & 97.10 &97.10&  100.0 \\
			&  T- Bow &82.66& \textbf{94.12} &97.15    & \textbf{76.92}  \\ 
			&   BoW &  \textbf{60.66}& 94.18 &\textbf{93.93}   & 99.99 \\ \hline

		\end{tabular}%
	}

\caption{{\bf Success rates of white-box attacks on the base network and on our BoW network.} Note that, whether attacked directly (BoW) or via the base network (T-BoW), our model is more robust to adversarial attacks.}  
\label{table:asr}
	
\end{table}

\subsubsection{Detecting Adversarial Samples}

\textbf{Adversarial detector.} 
We now evaluate the effectiveness of the detection strategy introduced in Section~\ref{sec:method_detection}. To this end, we train our detector using adversarial examples generated by the white-box attacks discussed above. Specifically, we report results obtained with two training strategies. {\it Strategy 1}, which is commonly used~\cite{ma2018characterizing}, consists of defining a balanced training set comprised of normal images (positives) and their adversarial counterparts (negatives). A drawback of this strategy, however, is that, as shown above, many attacks are unsuccessful with our BoW model, and considering such samples as negatives essentially adds noise to our training process, since unsuccessful adversarial samples will typically activate a codeword that is similar to the input image. To overcome this, we therefore propose {\it Strategy 2}, which consists of using only the successful adversarial examples as negatives during training. The resulting training set, however, is then imbalanced. The detailed detector architectures are provided in the supplementary material.

\vspace{0.1cm}
\noindent{\bf Comparison with the state of the art.} 
We compare our approach with the state-of-the-art BU~\cite{BU} and LID~\cite{ma2018characterizing} methods, which have proven more robust than the earlier detection strategy~\cite{KD}. 
In Table~\ref{tab:lid_BoW}, we report the area under the ROC curve (AUROC) for BU, LID and our method, using both Strategy 1 and Strategy 2. In this case, the results were obtained using the adversarial examples generated with a white-box attack on our BoW model directly. As such, BU and LID were also applied to our BoW model.
Note that we outperform the state of the art in all but one case, by a particularly large margin when the results are not already saturated, such as with BIM-a on SVHN and CIFAR-10 and with FGSM on CIFAR-10. The only case where we don't is with CW on MNIST, with Strategy 1. Note, however, that, as shown in Table~\ref{table:asr}, in this case, CW only has a 3.9\% success rate. This results in Strategy 1 having access to very few training samples.  By contrast, with Strategy 2, we outperform the baselines for the CW attack by a large margin.
In Table~\ref{tab:lid_base}, we provide similar results for attacks targeted to the base network. The conclusions are the same: We outperform the baselines in most cases, and by a large margin where there remained room for improvement.

Similarly to~\cite{ma2018characterizing}, we also evaluate whether our detector trained for one specific attack generalizes to other ones. \ms{In Table~\ref{table:adv_generalizibilty}, we compare the generalizability of our approach and of LID. For each method, we show the results of the model that was trained on the attack that makes it generalize best to the other ones. Note that the performance of our detector is virtually unaffected. While LID also generalizes well, our method still outperforms it in most cases, and by a large margin in several scenarios.}

\comment{
Similarly to~\cite{ma2018characterizing}, we also evaluate whether our detector trained for one specific attack generalizes to other ones. As shown in Table \ref{table:adv_generalizibilty}, training on either CW \MS{In the table, it says FGSM}\KN{ LID's FGSM generalizes well and for our approach BIM-a or CW generalizes well. For FMNIST, CW was used in our case } or BIM-a attacks leaves the performance of our detector on the other ones virtually unaffected. While LID also generalizes well \kn{with training on FGSM}, our method still outperforms it in most cases, and by a large margin in several scenarios.

Similarly to~\cite{ma2018characterizing}, we also evaluate whether our detector trained for one specific attack generalizes to other ones. As shown in Table \ref{table:adv_generalizibilty}, training on either CW \MS{In the table, it says FGSM}\KN{ LID's FGSM generalizes well and for our approach BIM-a or CW generalizes well. For FMNIST, CW was used in our case } or BIM-a attacks leaves the performance of our detector on the other ones virtually unaffected. While LID also generalizes well, our method still outperforms it in most cases, and by a large margin in several scenarios.
}
\comment{
Similarly to~\cite{ma2018characterizing}, we also evaluate whether our detector trained for one specific attack generalizes to other ones. \ms{In Table~\ref{table:adv_generalizibilty}, we compare the generalizability of our approach and of LID. 
Note that the performance of our detector is virtually unaffected. While LID also generalizes well, our method still outperforms it in most cases, and by a large margin in several scenarios.}}





\begin{table}[t]

	\resizebox{\linewidth}{!}{%
		\begin{tabular}{|c|c|c|c|c|c|}
			\hline
			
			\multirow{1}{*}{Dataset}  & Feature & FGSM & BIM-a & BIM-b  & CW  \\ \hline
			
			\multirow{3}{*}{MNIST} 
			& KD+BU  &  95.22/94.11& 82.54/72.80&82.17/72.43& 50.17/60.01\\
			&  LID  & 92.66/91.18 & 82.24/61.90 & 83.06/75.61 & \textbf{51.81}/68.46\\ 
			&  \textbf{Ours} &   \textbf{100.00/100.0}  & \textbf{100.0/100.0} & \textbf{100.0/100.0} &50.87/ \textbf{100.0} \\ \hline
			
			\multirow{3}{*}{F-MNIST} 
			& KD+BU    &99.33/99.36&91.35/87.32&89.43/85.39&69.68/65.83\\
			&  LID  & 93.88/93.93 & 86.95/81.06 & 86.83/81.11 & 73.23/70.84 \\ 
			& \textbf{ Ours} &   \textbf{100.0/100.0} & \textbf{100.0/100.0} & \textbf{100.0/100.0} &  \textbf{ 83.97/97.96} \\ \hline
			
			\multirow{3}{*}{SVHN} 
			& KD+BU    &78.21/74.24&78.37/73.97&67.96/71.46&88.68/88.62\\
			&  LID         &99.88/99.25&83.45/84.17& 85.76/90.42& 91.23/91.40 \\ 
			&  \textbf{Ours}      & \textbf{99.9/100.0} & \textbf{98.71/96.16} & \textbf{99.9/100.0} & \textbf{96.52/96.49}\\ \hline
			
			\multirow{3}{*}{CIFAR-10} 
			& KD+BU   & 72.79/69.66&86.23/85.69&60.23/62.52& 93.74/93.74\\
			&  LID  &89.67/89.26 & 85.40/85.02 & 80.55/82.79  & 93.57/93.17  \\ 
			&  \textbf{Ours} & \textbf{99.37/99.97}&\textbf{93.90/97.47}&\textbf{99.90/99.96}&\textbf{96.52/96.35}    \\ \hline
			
		\end{tabular}%
	}
	\caption{{\bf AUROC scores for BU, LID and our detector for direct BoW attacks.} Left and right numbers correspond to Strategy 1 and Strategy 2, respectively.} 
\label{tab:lid_BoW}
\vspace*{-0.5cm}
\end{table}


\begin{table}[t!]

	\resizebox{\linewidth}{!}{%
		\begin{tabular}{|c|c|c|c|c|c|c|}
			\hline
			
			\multirow{1}{*}{Dataset}  & Feature & FGSM & BIM-a & BIM-b  & CW  \\ \hline
			
			\multirow{3}{*}{MNIST} 
			& KD+BU    &93.73/93.63&86.09/83.09&79.22/79.22&80.64/80.64\\
			&  LID  & 99.17/99.15 & 99.75/99.75 & 95.16/96.16 & 99.01/99.03\\ 
			&  \textbf{Ours} &   \textbf{100.00/100.0}  & \textbf{100.0/100.0} & \textbf{100.0/100}.&\textbf{99.85/100.0} \\ \hline
			
			\multirow{3}{*}{F-MNIST} 
			& KD+BU  &  96.99/97.03&92.60/92.60&97.74/97.74&93.27/93.27\\
			&  LID  & 95.68/95.68 & 95.95/95.95 & 95.24/95.24 &  97.31/97.31 \\ 
			&  \textbf{Ours} &   \textbf{100.0/100.0} & \textbf{99.98/100.0} & \textbf{100.0/100.0} & \textbf{98.26}/\textbf{97.96} \\ \hline
			
			\multirow{3}{*}{SVHN} 
			& KD+BU    & 85.04/85.08&88.06/88.06&99.99/99.99&92.85/92.85\\
			&  LID  &99.88/99.32&88.45/84.17 & 98.76/99.93& 94.23/94.23 \\ 
			&  \textbf{Ours} &   \textbf{99.99}/\textbf{100.0} & \textbf{96.66/96.25} & \textbf{100.0/100.0} &   \textbf{95.61/96.86} \\ \hline
			
			\multirow{3}{*}{CIFAR-10} 
			& KD+BU    &75.75/74.67&80.02/79.69&99.12/99.24&96.06/96.05\\
			&  LID  & 87.16/87.92 & 82.17/81.77 & \textbf{99.91/99.91} &  \textbf{97.54}/\textbf{97.56} \\ 
			&  \textbf{Ours} &   \textbf{99.80}/\textbf{99.71} & \textbf{96.77}/\textbf{96.87} &99.77/99.85 & 96.70/97.24 \\ \hline
			
		\end{tabular}%
	}
	\caption{{\bf AUROC scores for BU, LID and our detector for attacks on the base network.} Left and right numbers correspond to Strategy 1 and Strategy 2, respectively.} 
\label{tab:lid_base}
\end{table}

\begin{table}[tt!]

	\resizebox{\linewidth}{!}{%
		
		\begin{tabular}{|c|c|c|cccc|}
			\hline
			\multirow{1}{*}{Dataset}  &Method & Train & FGSM & BIM-a & BIM-b&  CW  \\ \hline
			
			\multirow{2}{*}{MNIST} 
			& LID  & FGSM &91.18 &65.48&64.42& 29.05\\
			&   \textbf{Ours} & BIM-a &\textbf{100.0}&\textbf{100.0}&\textbf{100.0}&\textbf{97.56}\\ \hline
			
			\multirow{2}{*}{F-MNIST} 
			
			& LID  &FGSM &93.88&82.24&82.58&65.03\\
			&   \textbf{Ours} & CW &\textbf{97.39}&\textbf{97.13}&\textbf{95.85}&\textbf{97.96}\\ \hline
			
			\multirow{2}{*}{SVHN-10} 
			
			& LID  & FGSM &\textbf{99.25}&77.54&79.77&75.16\\
			&   \textbf{Ours} & BIM-a &91.41&\textbf{ 96.25} & \textbf{91.30} & \textbf{94.73}\\ \hline

			\multirow{2}{*}{CIFAR-10} 
			& LID & FGSM &\textbf{89.26}&66.55&68.33& 66.05\\
			&   \textbf{Ours} & BIM-a &86.90& \textbf{97.47}& \textbf{95.02}& \textbf{95.44} \\ \hline

		\end{tabular}%
	}
	
	\caption{{\bf Generalizing to different attacks.} We compare the results of LID and our detector in the scenario where the detectors were trained for a specific attack, but tested on different ones. These results were obtained with Strategy 2.}  
	\label{table:adv_generalizibilty}
	
\end{table}

\subsection{Attacking  the Detector}

In the previous set of experiments, we have worked under the assumption that the attacker only had access to our BoW model, but not to the detector. Here, we remove this assumption, and study two more challenging scenarios. In the first one, the attacker knows our detection strategy, but not the model we use. In the second, the attacker also has access to our detector.

\subsubsection{Adaptive Attack}

To evaluate the robustness of our approach in the scenario where the attacker is aware of our detection scheme, we apply an adaptive CW attack strategy similar to the ones used in~\cite{cwbypass,ma2018characterizing} to attack the KD \& LID detectors, respectively. To this end, we modify the objective of the CW attack as
\begin{equation}\label{eq:adaptive}
\arg \min_{\bI_{adv}}\left\|(\bI-\bI_{adv}\right\|_{2}^{2}+ \alpha \cdot \big(\ell(\bI_{adv})+\left\|\phi(\bI)-\phi(\bI_{adv})\right\|_{2}^{2}).
\end{equation}
The first two terms correspond to the original CW attack, with $\alpha$ balancing the amount of perturbation and the adversarial strength, that is, how strongly one forces the adversarial image to be misclassified. The last term directly reflects our detection strategy and encourages the BoW representation of the real, $\phi(\bI)$, and adversarial, $\phi(\bI_{adv})$, images to be similar. The rationale behind this is that the attack then aims to find an adversarial perturbation such that the sample is not only misclassified, but also has a representation close to that of the real image, thus breaking the premise on which our detector is built. \comment{The optimal constant $\alpha$ is determined via a search in the range $[10^{-3}, 10^{6}]$. }


\begin{table}[t!]    
		{\scriptsize 
	\resizebox{0.8\linewidth}{!}{%
		
		\begin{tabular}{|c|c|c|c|}
			\hline
			\multirow{1}{*}{Dataset}   &  Attack success rate &  Detector AUROC\\ \hline
			
			
			\multirow{1}{*}{F-MNIST}   
		 & 33.11 & 96.22  \\
			\hline
			
			\multirow{1}{*}{SVHN}   
		 & 94.58&  96.54   \\
			
			\hline
			\multirow{1}{*}{CIFAR-10}   
			 &  99.95 & 97.47     \\
			 \hline

		\end{tabular}%
	}
	}
\caption{{\bf Adaptive attacks.} Our detector remains robust to an attacker that knows our detection strategy.}
\label{table:CW_adaptive}
\end{table}

In Table~\ref{table:CW_adaptive}, we report both the success rate of this adaptive attack on our BoW model and the AUROC of our detector, trained with Strategy 2. Note that our detector still yields high AUROC, thus showing that it remains robust to an attacker that knows our detection strategy.


\subsubsection{White-box Detector Attacks}
We now evaluate the robustness of our method in the case where the attacker has access to all our models, that is, the BoW model, the GAN and the adversarial sample detector. Note that access to the parameters of the generator and detector networks is not a mild assumption since information about the training data is required to compute them.

To attack our complete framework, we generate adversarial images $\mathbf{\bI}_{adv}$ in a two-step fashion. First, we attack the BoW classifier to generate an intermediate adversarial image $\bI_{adv}^{0}$ that activates codeword $j$ corresponding to image $\bV_j$ in the visual codebook. Second, we attack the detector to misclassify $\bV_j$ as being similar to the input image. Since our detector relies on the distance between the input image and the codeword in feature space, fooling it can be achieved by finding a perturbation that solves the optimization problem
\begin{equation}
\min_{\bI_{adv}}\left\|\bI_{adv}^{0}-\bI_{adv}\right\|_{2}^{2}+\alpha \cdot\left\|\gamma(\bV_j)-\gamma(\bI_{adv})\right\|_{2}^{2}\;,
\label{eq:whitebox}
\end{equation}
where $\gamma(\cdot)$ is the feature-extraction part of our detector.

\begin{table}[t!]    
	
	\resizebox{\linewidth}{!}{%
		
		\begin{tabular}{|c|c|c|c|c|}
			\hline
			\multirow{2}{*}{Dataset}  & Classifier  & Detector & Attack success rate  & Attack success rate on detector \\ 
			&attack&attack &on detector& trained with adversarial training \\ \hline
			
			\vspace{0.1cm}
			
			\multirow{2}{*}{MNIST} 
			
			& FGSM &  FGSM & 0.00 &  0.0 \\
			& CW &  CW & 100.00 &  8.0 \\
			\hline
			
			\multirow{2}{*}{F-MNIST} 
			
			& FGSM &  FGSM & 2.8 &  0.0 \\
			& CW &  CW &  100 &  27.2 \\
			\hline
			
		\end{tabular}%
		
	}
\caption{{\bf White-box detector attacks.} An attacker that has access to our BoW network, GAN and detector can indeed be successful when using the CW attack, but not the FGSM one. However, adversarial training allows us to robustify our approach to these attacks, as shown in the right column. }
\label{table:CW_whitebox}
\end{table}

We found that, as is, our detector is indeed vulnerable to such an attack. To circumvent this, we therefore rely on the adversarial training strategy of~\cite{Adv_training}, in which the detector is trained using dynamically-generated adversarial images. We observed that, after a few epochs of such dynamic training, our detector becomes robust to these white-box attacks. To evidence this, in Table~\ref{table:CW_whitebox}, we provide the results of a white-box attack on a detector dynamically trained on limited balanced set of 10K samples and evaluated on 1K samples. These results show that our approach has become much more robust to this attack.


\subsection{Detecting Out-of-distribution Samples}

We now evaluate the use of our approach to detect OOD samples. Following the setup of~\cite{OODbaseline}, we perform experiments using either SVHN~\cite{netzer2011reading} or MNIST~\cite{cifar10} as training datasets from which the in-distribution samples are drawn. The goal then is to detect OOD samples coming from other datasets, such as LSUN~\cite{lsun}, TinyImageNet~\cite{tiny}, Omniglot~\cite{omniglot}, Not-MNIST~\cite{notMNIST}. We consider two settings: In the first, the detection method does not see {\it any} OOD samples during training, but has access to adversarial examples generated by the BIM-a attack; in the second, the detector has access to 1000 images from the OOD dataset.
In Table~\ref{table:OOD}, we compare the results of our approach with those of the baseline method~\cite{OODbaseline} and of ODIN~\cite{ODIN}. Note that we clearly outperform them, both when we see OOD samples during training and in the more realistic case where we don't. We believe that this demonstrates the generality of our approach.
 
\comment{

\begin{table}[t!]
	
	\resizebox{0.95\linewidth}{!}{%
		
		\begin{tabular}{|c|c|c|c|}
			\hline
			\multirow{2}{*}{In-Dataset}  &  \multirow{2}{*}{In-Dataset}  &  Adversarial examples  & OOD samples \\
			&    & \multicolumn{2}{c|}{Baseline~\cite{OODbaseline} / ODIN~\cite{ODIN} / \textbf{Ours}}  \\
			\hline
			\multirow{2}{*}{SVHN} 
			& CIFAR-10 &       87.44/87.54/\textbf{97.26} &   87.92/87.46/\textbf{99.14}\\
			& LSUN    &   89.06/89.06/\textbf{99.87} & 89.06/89.52/\textbf{99.98}\\
			& TinyImageNet &89.97/90.42/\textbf{99.76}&   89.97/88.96/\textbf{99.93}\\
			
			\hline
			
			\multirow{2}{*}{MNIST-10} 
			& Not-MNIST    &       77.11/77.69/\textbf{97.44} &   77.23/77.69/\textbf{99.98}\\
			& OMNIGLOT    &  82.11/82.06/\textbf{97.01}&    82.24/82.06/\textbf{100.0}\\
			& CIFAR    & 79.84/79.77/\textbf{99.21}&  79.84/79.77/\textbf{100.0}\\

			\hline
		\end{tabular}%
	}
\caption{{\bf{Detecting OOD samples.}} We compare our detector with the baseline method of~\cite{OODbaseline} and with the recent ODIN~\cite{ODIN} one in the case where we only observe adversarial samples during training (left) and when we have access to a few OOD samples (right).} \label{table:OOD}
	
\vspace*{-0.3cm}	
	
\end{table}

}

\begin{table}[t!]
	
	\resizebox{0.95\linewidth}{!}{%
		
		\begin{tabular}{|c|c|c|c|}
			\hline
			\multirow{2}{*}{In-Dataset}  &  \multirow{2}{*}{In-Dataset}  &  Adversarial examples  & OOD samples \\
			&    & \multicolumn{2}{c|}{Baseline~\cite{OODbaseline}  / \textbf{Ours}}  \\
			\hline
			\multirow{2}{*}{SVHN} 
			& CIFAR-10 &       87.44/\textbf{97.26} &   87.92/\textbf{99.14}\\
			& LSUN    &   89.06/\textbf{99.87} & 89.06/\textbf{99.98}\\
			& TinyImageNet &89.97/\textbf{99.76}&   89.97/\textbf{99.93}\\
			
			\hline
			
			\multirow{2}{*}{MNIST-10} 
			& Not-MNIST    &       77.11/\textbf{97.44} &   77.23/\textbf{99.98}\\
			& OMNIGLOT    &  82.11/\textbf{97.01}&    82.24/\textbf{100.0}\\
			& CIFAR    & 79.84/\textbf{99.21}&  79.84/\textbf{100.0}\\

			\hline
		\end{tabular}%
	}
	\caption{{\bf{Detecting OOD samples.}} We compare our detector with the baseline method of~\cite{OODbaseline} and with the recent ODIN~\cite{ODIN} one in the case where we only observe adversarial samples during training (left) and when we have access to a few OOD samples (right).} \label{table:OOD}
	
	\vspace*{-0.3cm}	
	
\end{table}

\section{Conclusion}
We have introduced a novel approach to interpreting a CNN's prediction, by providing the elements in a BoW codebook with a visual and semantic meaning. We have then proposed to leverage the visual representation of these interpretable BoW networks for adversarial example detection. Our experiments have evidenced that (i) our interpretable BoW networks are more robust to adversarial attacks; (ii) our adversary detection strategy outperforms the state-of-the-art ones; (iii) our approach could be made robust to adversarial attacks to the detector itself; (iv) our framework generalizes to OOD sample detection. In the future, following the intuition that complex scenes and objects can be modeled as sets of parts, we plan to extend our interpretable BoW representation to part-based ones. This, we believe, will also translate to improved adversary detection performance, since it will allow us to robustly combine the decision of each part.

\section{Conclusion}
We have introduced a novel approach to interpreting a CNN's prediction, by providing the elements in a BoW codebook with a visual and semantic meaning. We have then proposed to leverage the visual representation of these interpretable BoW networks for adversarial example detection. Our experiments have evidenced that (i) our interpretable BoW networks are more robust to adversarial attacks; (ii) our adversary detection strategy outperforms the state-of-the-art ones; (iii) our approach could be made robust to adversarial attacks to the detector itself; (iv) our framework generalizes to OOD sample detection. In the future, following the intuition that complex scenes and objects can be modeled as sets of parts, we plan to extend our interpretable BoW representation to part-based ones. This, we believe, will also translate to improved adversary detection performance, since it will allow us to robustly combine the decision of each part.


{\small
\bibliographystyle{ieee}
\bibliography{egbib}

\begin{thebibliography}{10}\itemsep=-1pt

\bibitem{netvlad2016}
R.~Arandjelovic, P.~Gronat, A.~Torii, T.~Pajdla, and J.~Sivic.
\newblock Netvlad: Cnn architecture for weakly supervised place recognition.
\newblock In {\em Proceedings of the IEEE conference on computer vision and
  pattern recognition}, pages 5297--5307, 2016.

\bibitem{allaboutvlad}
R.~Arandjelovic and A.~Zisserman.
\newblock All about vlad.
\newblock In {\em Proceedings of the IEEE conference on Computer Vision and
  Pattern Recognition}, pages 1578--1585, 2013.

\bibitem{notMNIST}
Y.~Bulatov.
\newblock notmnist dataset.
\newblock 2011.

\bibitem{cwbypass}
N.~Carlini and D.~Wagner.
\newblock Adversarial examples are not easily detected: Bypassing ten detection
  methods.
\newblock In {\em Proceedings of the 10th ACM Workshop on Artificial
  Intelligence and Security}, pages 3--14. ACM, 2017.

\bibitem{CWattack}
N.~Carlini and D.~Wagner.
\newblock Towards evaluating the robustness of neural networks.
\newblock In {\em 2017 IEEE Symposium on Security and Privacy (SP)}, pages
  39--57. IEEE, 2017.

\bibitem{KD}
R.~Feinman, R.~R. Curtin, S.~Shintre, and A.~B. Gardner.
\newblock Detecting adversarial samples from artifacts.
\newblock {\em arXiv preprint arXiv:1703.00410}, 2017.

\bibitem{felz_object2008}
P.~Felzenszwalb, D.~McAllester, and D.~Ramanan.
\newblock A discriminatively trained, multiscale, deformable part model.
\newblock In {\em Computer Vision and Pattern Recognition, 2008. CVPR 2008.
  IEEE Conference on}, pages 1--8. IEEE, 2008.

\bibitem{felz_object2010}
P.~F. Felzenszwalb, R.~B. Girshick, D.~McAllester, and D.~Ramanan.
\newblock Object detection with discriminatively trained part-based models.
\newblock {\em IEEE transactions on pattern analysis and machine intelligence},
  32(9):1627--1645, 2010.

\bibitem{felz_pictorial}
P.~F. Felzenszwalb and D.~P. Huttenlocher.
\newblock Pictorial structures for object recognition.
\newblock {\em International journal of computer vision}, 61(1):55--79, 2005.

\bibitem{attpool1}
R.~Girdhar and D.~Ramanan.
\newblock Attentional pooling for action recognition.
\newblock In {\em Advances in Neural Information Processing Systems}, pages
  34--45, 2017.

\bibitem{actionvlad}
R.~Girdhar, D.~Ramanan, A.~Gupta, J.~Sivic, and B.~Russell.
\newblock Actionvlad: Learning spatio-temporal aggregation for action
  classification.
\newblock In {\em CVPR}, volume~2, page~3, 2017.

\bibitem{multiscale}
Y.~Gong, L.~Wang, R.~Guo, and S.~Lazebnik.
\newblock Multi-scale orderless pooling of deep convolutional activation
  features.
\newblock In {\em European conference on computer vision}, pages 392--407.
  Springer, 2014.

\bibitem{gan2014}
I.~Goodfellow, J.~Pouget-Abadie, M.~Mirza, B.~Xu, D.~Warde-Farley, S.~Ozair,
  A.~Courville, and Y.~Bengio.
\newblock Generative adversarial nets.
\newblock In {\em Advances in neural information processing systems}, pages
  2672--2680, 2014.

\bibitem{fgsm}
I.~J. Goodfellow, J.~Shlens, and C.~Szegedy.
\newblock Explaining and harnessing adversarial examples (2014).
\newblock {\em arXiv preprint arXiv:1412.6572}.

\bibitem{Adv_training}
I.~J. Goodfellow, J.~Shlens, and C.~Szegedy.
\newblock Explaining and harnessing adversarial examples (2014).
\newblock {\em arXiv preprint arXiv:1412.6572}.

\bibitem{grosse2017}
K.~Grosse, P.~Manoharan, N.~Papernot, M.~Backes, and P.~McDaniel.
\newblock On the (statistical) detection of adversarial examples.
\newblock {\em arXiv preprint arXiv:1702.06280}, 2017.

\bibitem{BU}
K.~Grosse, P.~Manoharan, N.~Papernot, M.~Backes, and P.~McDaniel.
\newblock On the (statistical) detection of adversarial examples.
\newblock {\em arXiv preprint arXiv:1702.06280}, 2017.

\bibitem{WGAN-GP}
I.~Gulrajani, F.~Ahmed, M.~Arjovsky, V.~Dumoulin, and A.~C. Courville.
\newblock Improved training of wasserstein gans.
\newblock In {\em Advances in Neural Information Processing Systems}, pages
  5767--5777, 2017.

\bibitem{CLoss}
R.~Hadsell, S.~Chopra, and Y.~LeCun.
\newblock Dimensionality reduction by learning an invariant mapping.
\newblock In {\em null}, pages 1735--1742. IEEE, 2006.

\bibitem{resnet}
K.~He, X.~Zhang, S.~Ren, and J.~Sun.
\newblock Deep residual learning for image recognition.
\newblock In {\em Proceedings of the IEEE conference on computer vision and
  pattern recognition}, pages 770--778, 2016.

\bibitem{OODbaseline}
D.~Hendrycks and K.~Gimpel.
\newblock A baseline for detecting misclassified and out-of-distribution
  examples in neural networks.
\newblock {\em arXiv preprint arXiv:1610.02136}, 2016.

\bibitem{vlad2012}
H.~Jegou, F.~Perronnin, M.~Douze, J.~S{\'a}nchez, P.~Perez, and C.~Schmid.
\newblock Aggregating local image descriptors into compact codes.
\newblock {\em IEEE transactions on pattern analysis and machine intelligence},
  34(9):1704--1716, 2012.

\bibitem{adam}
D.~P. Kingma and J.~Ba.
\newblock Adam: A method for stochastic optimization.
\newblock {\em arXiv preprint arXiv:1412.6980}, 2014.

\bibitem{bow1}
J.~J. Koenderink and A.~J. Van~Doorn.
\newblock The structure of locally orderless images.
\newblock {\em International Journal of Computer Vision}, 31(2-3):159--168,
  1999.

\bibitem{cifar10}
A.~Krizhevsky and G.~Hinton.
\newblock Learning multiple layers of features from tiny images.
\newblock Technical report, Citeseer, 2009.

\bibitem{tiny}
A.~Krizhevsky and G.~Hinton.
\newblock Learning multiple layers of features from tiny images.
\newblock Technical report, Citeseer, 2009.

\bibitem{alexnet}
A.~Krizhevsky, I.~Sutskever, and G.~E. Hinton.
\newblock Imagenet classification with deep convolutional neural networks.
\newblock In {\em Advances in neural information processing systems}, pages
  1097--1105, 2012.

\bibitem{bima}
A.~Kurakin, I.~Goodfellow, and S.~Bengio.
\newblock Adversarial examples in the physical world.
\newblock {\em arXiv preprint arXiv:1607.02533}, 2016.

\bibitem{omniglot}
B.~M. Lake, R.~Salakhutdinov, and J.~B. Tenenbaum.
\newblock Human-level concept learning through probabilistic program induction.
\newblock {\em Science}, 350(6266):1332--1338, 2015.

\bibitem{mnist}
Y.~LeCun, L.~Bottou, Y.~Bengio, and P.~Haffner.
\newblock Gradient-based learning applied to document recognition.
\newblock {\em Proceedings of the IEEE}, 86(11):2278--2324, 1998.

\bibitem{Mahalanobis2018}
K.~Lee, K.~Lee, H.~Lee, and J.~Shin.
\newblock A simple unified framework for detecting out-of-distribution samples
  and adversarial attacks.
\newblock {\em arXiv preprint arXiv:1807.03888}, 2018.

\bibitem{objectbanks}
L.-J. Li, H.~Su, L.~Fei-Fei, and E.~P. Xing.
\newblock Object bank: A high-level image representation for scene
  classification \& semantic feature sparsification.
\newblock In {\em Advances in neural information processing systems}, pages
  1378--1386, 2010.

\bibitem{ODIN}
S.~Liang, Y.~Li, and R.~Srikant.
\newblock Enhancing the reliability of out-of-distribution image detection in
  neural networks.
\newblock {\em arXiv preprint arXiv:1706.02690}, 2017.

\bibitem{ma2018characterizing}
X.~Ma, B.~Li, Y.~Wang, S.~M. Erfani, S.~Wijewickrema, M.~E. Houle,
  G.~Schoenebeck, D.~Song, and J.~Bailey.
\newblock Characterizing adversarial subspaces using local intrinsic
  dimensionality.
\newblock {\em arXiv preprint arXiv:1801.02613}, 2018.

\bibitem{visualizing2}
A.~Mahendran and A.~Vedaldi.
\newblock Understanding deep image representations by inverting them.
\newblock In {\em Proceedings of the IEEE conference on computer vision and
  pattern recognition}, pages 5188--5196, 2015.

\bibitem{visualizing1}
A.~Mahendran and A.~Vedaldi.
\newblock Visualizing deep convolutional neural networks using natural
  pre-images.
\newblock {\em International Journal of Computer Vision}, 120(3):233--255,
  2016.

\bibitem{metzen2017}
J.~H. Metzen, T.~Genewein, V.~Fischer, and B.~Bischoff.
\newblock On detecting adversarial perturbations.
\newblock {\em arXiv preprint arXiv:1702.04267}, 2017.

\bibitem{DF}
S.-M. Moosavi-Dezfooli, A.~Fawzi, and P.~Frossard.
\newblock Deepfool: a simple and accurate method to fool deep neural networks.
\newblock In {\em Proceedings of the IEEE Conference on Computer Vision and
  Pattern Recognition}, pages 2574--2582, 2016.

\bibitem{nakka2018}
K.~K. Nakka and M.~Salzmann.
\newblock Deep attentional structured representation learning for visual
  recognition.
\newblock {\em BMVC 2018}, 2018.

\bibitem{netzer2011reading}
Y.~Netzer, T.~Wang, A.~Coates, A.~Bissacco, B.~Wu, and A.~Y. Ng.
\newblock Reading digits in natural images with unsupervised feature learning.
\newblock In {\em NIPS workshop on deep learning and unsupervised feature
  learning}, volume 2011, page~5, 2011.

\bibitem{visualizing3}
A.~Nguyen, A.~Dosovitskiy, J.~Yosinski, T.~Brox, and J.~Clune.
\newblock Synthesizing the preferred inputs for neurons in neural networks via
  deep generator networks.
\newblock In {\em Advances in Neural Information Processing Systems}, pages
  3387--3395, 2016.

\bibitem{fisher2}
F.~Perronnin, J.~S{\'a}nchez, and T.~Mensink.
\newblock Improving the fisher kernel for large-scale image classification.
\newblock In {\em European conference on computer vision}, pages 143--156.
  Springer, 2010.

\bibitem{bow2}
P.~Quelhas, F.~Monay, J.-M. Odobez, D.~Gatica-Perez, T.~Tuytelaars, and
  L.~Van~Gool.
\newblock Modeling scenes with local descriptors and latent aspects.
\newblock In {\em Computer Vision, 2005. ICCV 2005. Tenth IEEE International
  Conference on}, volume~1, pages 883--890. IEEE, 2005.

\bibitem{dcgan2015}
A.~Radford, L.~Metz, and S.~Chintala.
\newblock Unsupervised representation learning with deep convolutional
  generative adversarial networks.
\newblock {\em arXiv preprint arXiv:1511.06434}, 2015.

\bibitem{fisher1}
J.~S{\'a}nchez, F.~Perronnin, T.~Mensink, and J.~Verbeek.
\newblock Image classification with the fisher vector: Theory and practice.
\newblock {\em International journal of computer vision}, 105(3):222--245,
  2013.

\bibitem{gradcam}
R.~R. Selvaraju, M.~Cogswell, A.~Das, R.~Vedantam, D.~Parikh, D.~Batra, et~al.
\newblock Grad-cam: Visual explanations from deep networks via gradient-based
  localization.
\newblock In {\em ICCV}, pages 618--626, 2017.

\bibitem{bow3}
J.~Sivic and A.~Zisserman.
\newblock Video google: A text retrieval approach to object matching in videos.
\newblock In {\em null}, page 1470. IEEE, 2003.

\bibitem{adv2013}
C.~Szegedy, W.~Zaremba, I.~Sutskever, J.~Bruna, D.~Erhan, I.~Goodfellow, and
  R.~Fergus.
\newblock Intriguing properties of neural networks.
\newblock {\em arXiv preprint arXiv:1312.6199}, 2013.

\bibitem{fishernet}
P.~Tang, X.~Wang, B.~Shi, X.~Bai, W.~Liu, and Z.~Tu.
\newblock Deep fishernet for object classification.
\newblock {\em arXiv preprint arXiv:1608.00182}, 2016.

\bibitem{classesmes}
L.~Torresani, M.~Szummer, and A.~Fitzgibbon.
\newblock Efficient object category recognition using classemes.
\newblock In {\em European conference on computer vision}, pages 776--789.
  Springer, 2010.

\bibitem{residualattention}
F.~Wang, M.~Jiang, C.~Qian, S.~Yang, C.~Li, H.~Zhang, X.~Wang, and X.~Tang.
\newblock Residual attention network for image classification.
\newblock {\em arXiv preprint arXiv:1704.06904}, 2017.

\bibitem{histnet2018}
Z.~Wang, H.~Li, W.~Ouyang, and X.~Wang.
\newblock Learnable histogram: Statistical context features for deep neural
  networks.
\newblock In {\em European Conference on Computer Vision}, pages 246--262.
  Springer, 2016.

\bibitem{fmnist}
H.~Xiao, K.~Rasul, and R.~Vollgraf.
\newblock Fashion-mnist: a novel image dataset for benchmarking machine
  learning algorithms.
\newblock {\em arXiv preprint arXiv:1708.07747}, 2017.

\bibitem{lsun}
J.~Xiao, J.~Hays, K.~A. Ehinger, A.~Oliva, and A.~Torralba.
\newblock Sun database: Large-scale scene recognition from abbey to zoo.
\newblock In {\em Computer vision and pattern recognition (CVPR), 2010 IEEE
  conference on}, pages 3485--3492. IEEE, 2010.

\bibitem{visualizing4}
M.~D. Zeiler and R.~Fergus.
\newblock Visualizing and understanding convolutional networks.
\newblock In {\em European conference on computer vision}, pages 818--833.
  Springer, 2014.

\bibitem{interpretableCNN}
Q.~Zhang, Y.~N. Wu, and S.-C. Zhu.
\newblock Interpretable convolutional neural networks.
\newblock {\em arXiv preprint arXiv:1710.00935}, 2(3):5, 2017.

\bibitem{CAM}
B.~Zhou, A.~Khosla, A.~Lapedriza, A.~Oliva, and A.~Torralba.
\newblock Learning deep features for discriminative localization.
\newblock In {\em Proceedings of the IEEE Conference on Computer Vision and
  Pattern Recognition}, pages 2921--2929, 2016.

\end{thebibliography}
}

\clearpage

Below, we provide additional details regarding our architectures and our experimental results.

\section{Architectures}
\label{sec:Supplementary}
As base networks for our models, we used the same architectures as in~\cite{ma2018characterizing}. These architectures are provided in Table~\ref{table:mnist_model} for the MNIST, FMNIST  experiments, in Table~\ref{table:svhn_model} for SVHN experiments and in Table~\ref{table:cifar_model} for the CIFAR-10 experiments. For the detector in our adversarial example detection approach, we used the architecture shown in Table~\ref{table:detector_mnist} for all datasets, except for CIFAR-10. In this case, we used a ResNet-20~\cite{resnet}\footnote{https://github.com/tensorflow/models/tree/master/official/resnet} 
due to the higher image variance and background complexity of this dataset.

We also trained a GAN~\cite{gan2014} for MNIST and FMNIST and a WGAN~\cite{WGAN-GP} for SVHN and CIFAR-10.  The architectures of  the generator for MNIST and FMNIST are provided in Table~\ref{table:generatpr_mnist} and in Table~\ref{table:generator_cifar} for SVHN and CIFAR-10. Similarly,  the discriminator  for MNIST and FMNIST is provided in Table~\ref{table:discriminator_mnist} and the one for SVHN and CIFAR-10 in Table~\ref{table:discriminator_cifar}.
\begin{table}[h]
	\centering
	\begin{tabular}{|c|c|}
		\hline
		Layer & Parameters  \\
		\hline
		Convolution + ReLU  & $5 \times 5 \times 64 $ \\
		Convolution + ReLU   & $5 \times 5 \times 64$ \\
		MaxPool & $2 \times 2$ \\
		Dense + ReLU  & 128\\
		Softmax & 10\\
		\hline
	\end{tabular}
	\caption{\textbf{Base network} for MNIST, FMNIST.}
	\label{table:mnist_model}
		\vspace*{-0.3cm}
\end{table}

\begin{table}[h]
	\centering
	\begin{tabular}{|c|c|}
		\hline
		Layer & Parameters  \\
		\hline
		Convolution + ReLU  & $3 \times 3 \times 64$ \\
		Convolution + ReLU   & $3 \times 3 \times 64$ \\
		MaxPool & $2 \times 2$ \\
		Convolution + ReLU  & $3 \times 3 \times 128$ \\
		Convolution + ReLU   & $3 \times 3 \times 128$ \\
		MaxPool & $2 \times 2$ \\
		Dense + ReLU  & 512\\
		Dense + ReLU  & 128\\
		Softmax & 10\\
		\hline
	\end{tabular}
	\caption{\textbf{Base network} for SVHN.}
	\label{table:svhn_model}
\vspace*{-0.3cm}
\end{table}

\begin{table}[h]
	\centering
	\begin{tabular}{|c|c|}
		\hline
		Layer & Parameters  \\
		\hline
		Convolution + ReLU  & $3 \times 3 \times 32$ \\
		Convolution + ReLU   & $3 \times 3 \times 32$ \\
		MaxPool & $2 \times 2$ \\
		Convolution + ReLU  & $3 \times 3 \times 64$ \\
		Convolution + ReLU   & $3 \times 3 \times 64$ \\
		MaxPool & $2 \times 2$ \\
		
		Convolution + ReLU  & $3 \times 3 \times 128$ \\
		Convolution + ReLU   & $3 \times 3 \times 128$ \\
		MaxPool & $2 \times 2$ \\
		Dense + ReLU  & 1024\\
		Dense + ReLU  & 512\\
		Softmax & 10\\
		\hline
	\end{tabular}
	\caption{\textbf{Base network} for CIFAR-10.}
	\label{table:cifar_model}
		\vspace*{-0.3cm}
\end{table}

\begin{table}[h]
	\centering
	\begin{tabular}{|c|c|}
		\hline
		Layer & Parameters  \\
		\hline
		Convolution + ReLU  & $5 \times 5 \times 32$ \\
		MaxPool & $2 \times 2$ \\
		Convolution + ReLU   & $5 \times 5 \times 64$ \\
		MaxPool & $2 \times 2$ \\
		Fully Connected + ReLU  & 128\\
		Fully Connected + ReLU  & 128\\
		\hline
	\end{tabular}
	\caption{\textbf{Detector} for MNIST, FMNIST and SVHN.}
	\label{table:detector_mnist}
		\vspace*{-0.3cm}
\end{table}

\begin{table}[h]
	\centering
	\begin{tabular}{|c|c|}
		\hline
		Layer & Parameters  \\ \hline 
		Input noise, $\bz \in \mathbb{R}^{128}$  &  Nil \\ 
		Dense + ReLU &  $128\times4\times4$ \\
		Deconvolution + ReLU  & $5 \times 5 \times 128$ \\
		Deconvolution + ReLU   & $5 \times 5 \times 64$ \\
		Deconvolution + Sigmoid   & $5 \times 5 \times 1$ \\
		\hline
	\end{tabular}
	\caption{\textbf{Generator } for MNIST and FMNIST}
	\label{table:generatpr_mnist}
		\vspace*{-0.3cm}
\end{table}

\begin{table}[h]
	\centering
	\begin{tabular}{|c|c|}
			\hline
		Layer & Parameters  \\
		\hline
		
				Input image, $\bI  \in \mathbb{R}^{28 \times 28}$ & Nil \\
		Convolution + LeakyReLU, stride=2  & $5 \times 5 \times 64$ \\
		Convolution + LeakyReLU, stride=2  & $5 \times 5 \times 128$ \\
		Convolution + LeakyReLU, stride=2  & $5 \times 5 \times 256$ \\
		Fully Connected + ReLU  & 1\\
		\hline
	\end{tabular}
	\caption{\textbf{Discriminator } for MNIST and FMNIST}
	\label{table:discriminator_mnist}
		\vspace*{-0.3cm}
\end{table}

\begin{table}[h]
	\centering
	\begin{tabular}{|c|c|}
		\hline
		Layer & Parameters  \\ \hline 
		Input noise, $\bz \in \mathbb{R}^{128}$  &  Nil \\ 
		Dense  &  $128\times4\times4$ \\
		ResBlock + Up  & $ 128$ \\
		ResBlock + Up  & $ 128$ \\
		ResBlock + Up  & $ 128$ \\
		Convolution + Tanh   & $3 \times 3 \times 3 $ \\
		\hline
	\end{tabular}
	\caption{\textbf{Generator } for SVHN and CIFAR-10 similar to the ones used in ~\cite{WGAN-GP}}
	\label{table:generator_cifar}
	\vspace*{-0.3cm}
\end{table}

\begin{table}[h]
	\centering
	\begin{tabular}{|c|c|}
		\hline
		Layer & Parameters  \\
		\hline
		
		Input image, $\bI  \in \mathbb{R}^{32 \times 32 \times 3}$ & Nil \\
		ResBlock + Down  & $ 128$ \\
		ResBlock + Down  & $ 128$ \\
		ResBlock +Down + GAP & $ 128$ \\
		Dense   & 1 \\
		\hline
	\end{tabular}
	\caption{\textbf{Discriminator } for SVHN and CIFAR-10 similar to the ones used in ~\cite{WGAN-GP}}
	\label{table:discriminator_cifar}
\end{table}

\section{Additional Results on Normal Samples}
\begin{figure*} [h] 
	\centering
	\vspace{-0.1in}
	\subfigure[MNIST]
	{
		\includegraphics[width=0.45\linewidth]{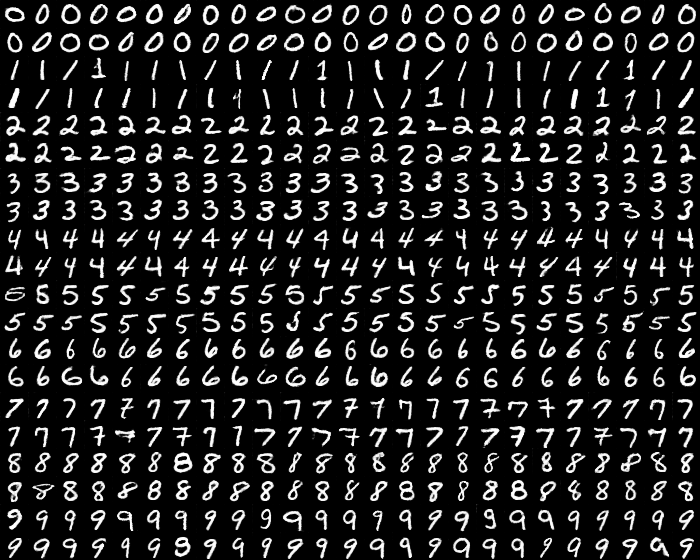} } 
	\,
	\subfigure[FMNIST]
	{
		\includegraphics[width=0.45\linewidth]{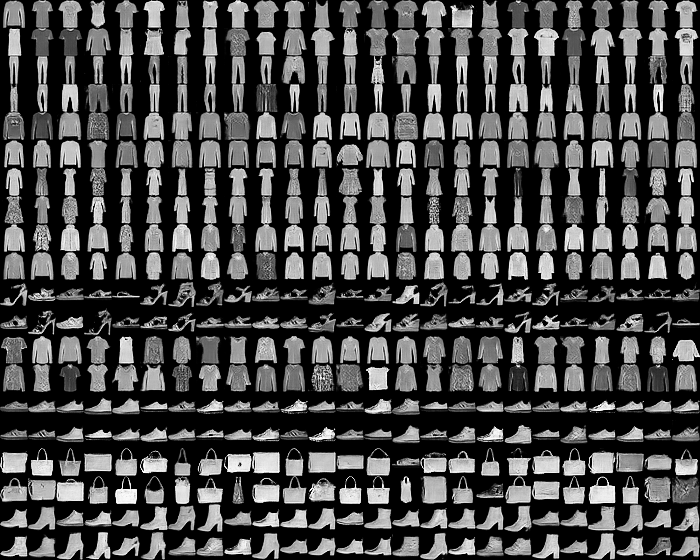} }
	\vspace{0.1cm}
	\subfigure[SVHN]
	{
		\includegraphics[width=0.45\linewidth]{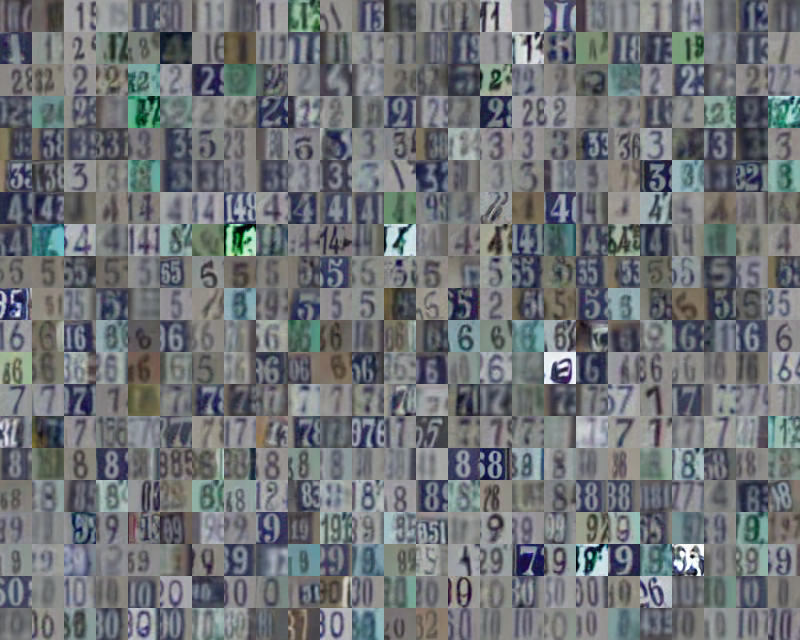}}
	\,
	\subfigure[CIFAR-10]
	{
		\includegraphics[width=0.45\linewidth]{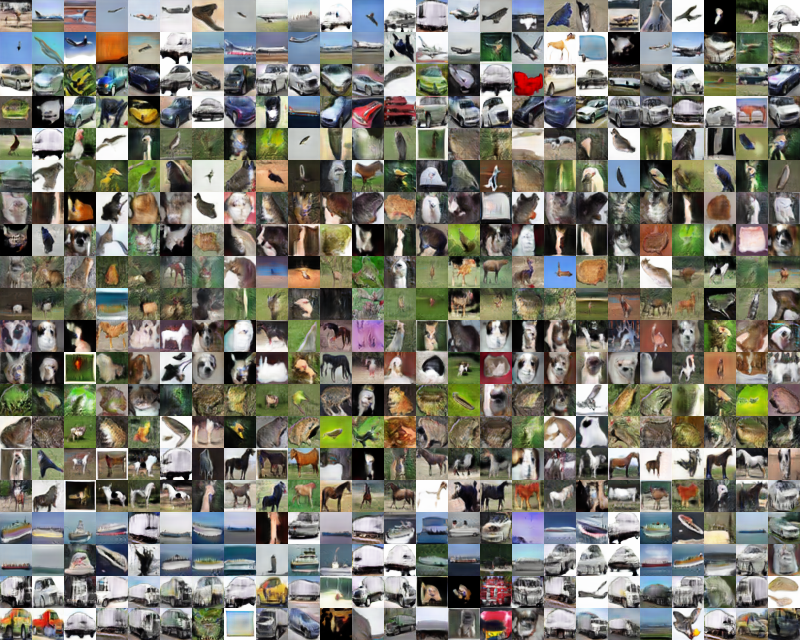}
	}
	
	\	\caption{{\bf Visual interpretations of the BoW codebooks.} Every two rows correspond to one class. Note that these codewords capture the diversity of the classes in each dataset (better seen by zooming in.)}
	\label{fig:codewords_full}
\end{figure*}

\begin{figure*} [h] \centering
	\vspace{-0.1in}
	\subfigure[MNIST]
	{
		\includegraphics[width=0.45\linewidth]{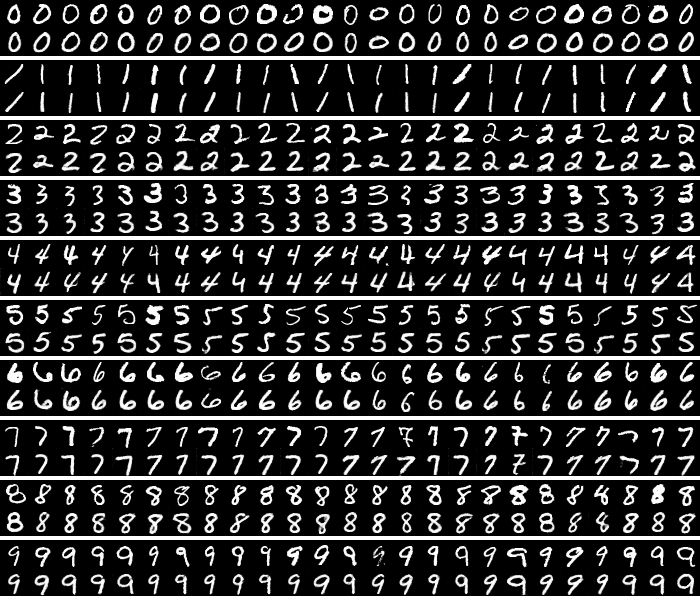} } 
	\,
	\subfigure[FMNIST]
	{
		\includegraphics[width=0.45\linewidth]{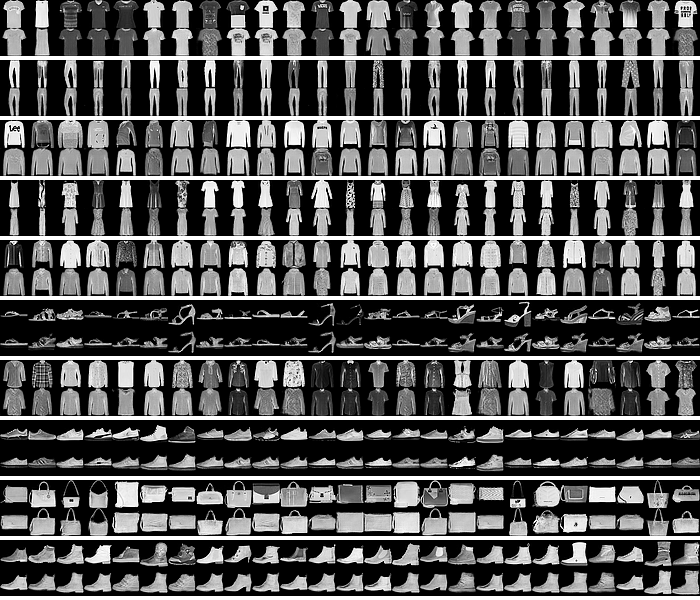} }
	\vspace{0.1cm}
	\subfigure[SVHN]
	{
		\includegraphics[width=0.45\linewidth]{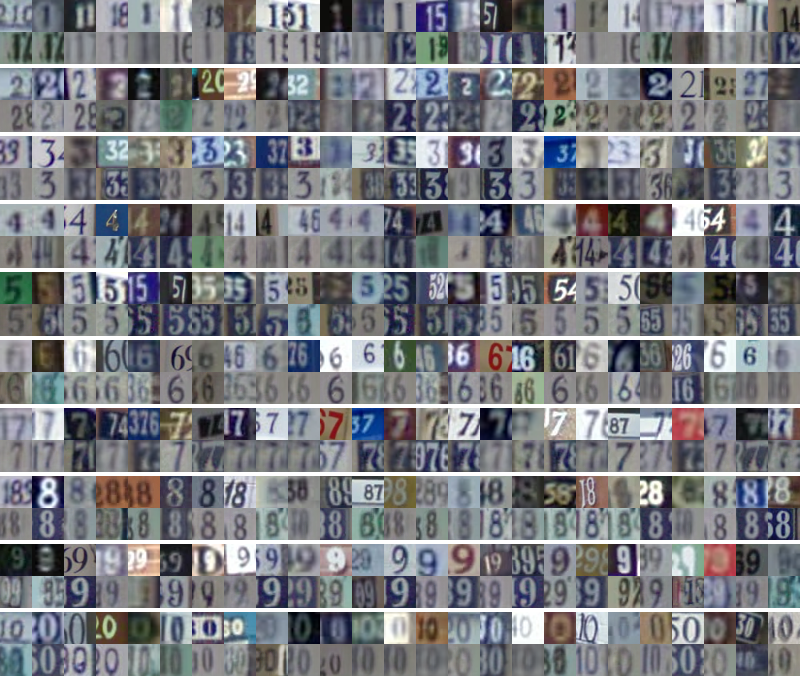}}
	\,
	\subfigure[CIFAR-10]
	{
		\includegraphics[width=0.45\linewidth]{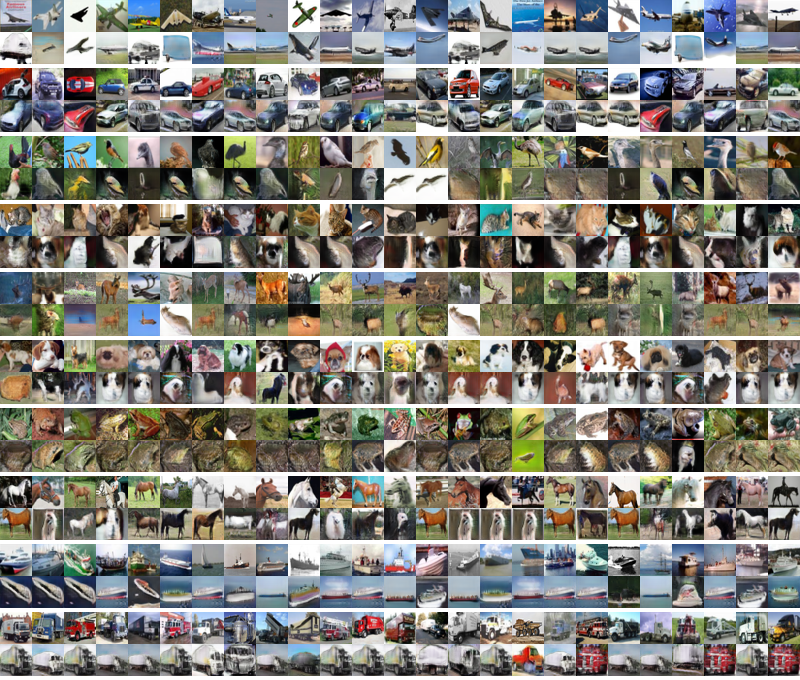}
	}
	
	\caption{{\bf Visualization of the most highly activated codeword for normal  test samples.} Each two rows within a block show the input image (top) and corresponding codeword (bottom). }\label{fig:normal_all}
	\vspace{+0.05in}
	
\end{figure*}

\begin{figure*} [h] \centering
	
	\includegraphics[width=0.95\linewidth]{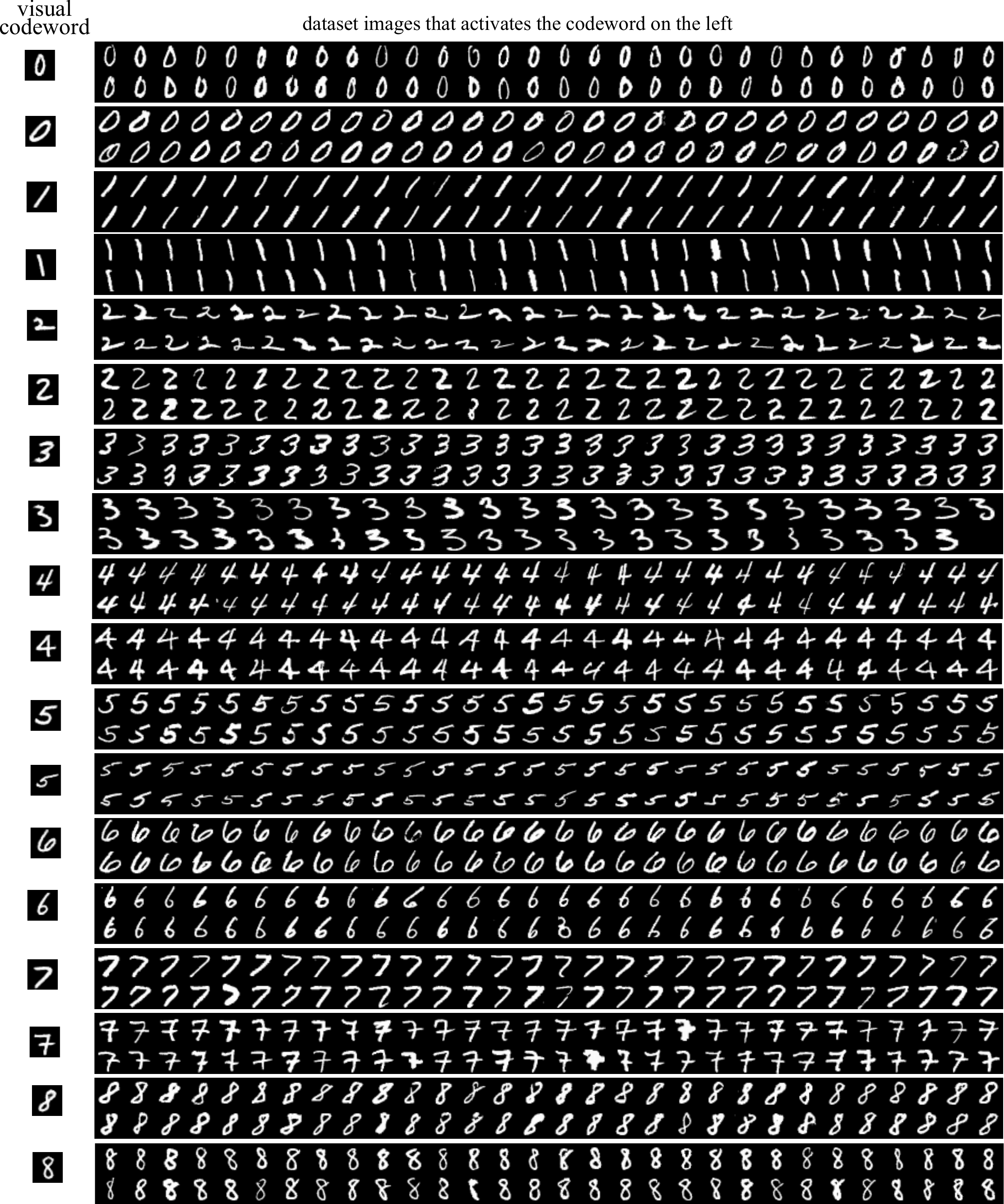} 
	
	\caption{{\bf Dataset images from MNIST that activate a particular codeword.} Each block has two columns. The left one shows the visual codeword and the right one  the test images that activated the codeword on the left.  Note that there is a clear semantic similarity between the images and the codeword, not only in terms of class label, but also with respect to style and orientation.}\label{fig:assgn_mnist}
	\vspace{+0.05in}
\end{figure*}

\begin{figure*} [h] \centering
	
	\includegraphics[width=0.95\linewidth]{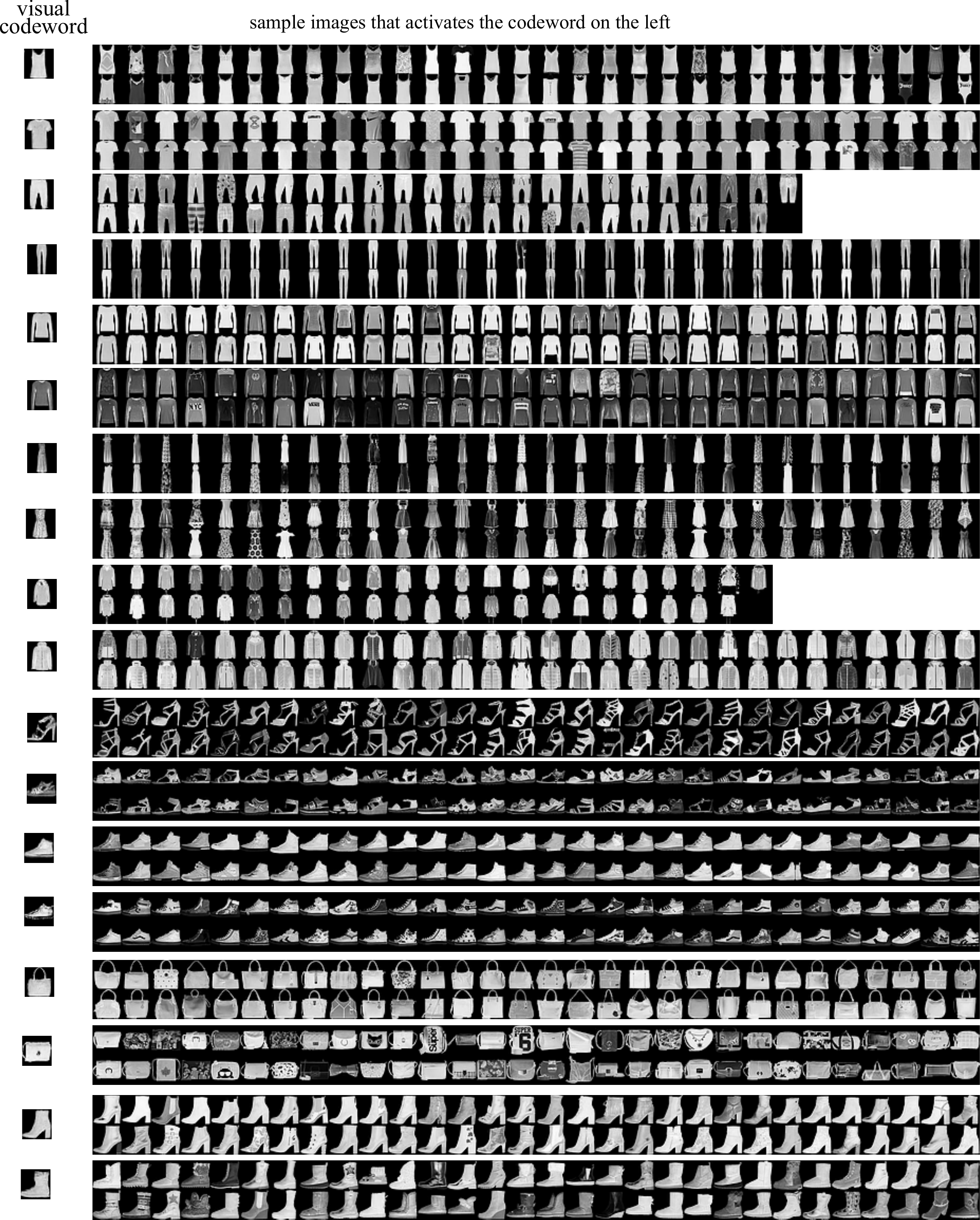} 
	
	\caption{{\bf  Dataset images from FMNIST that activate a particular codeword.} Each block has two columns. The left one shows the visual codeword and the right one  the dataset images that activated the codeword on the left.  Note that there is a clear semantic similarity between the images and the codeword, not only in terms of class label, but also with respect to style and orientation.}
	\vspace{+0.05in}\label{fig:assgn_fmnist}
	
\end{figure*}

\begin{figure*} [h] \centering
	
	\includegraphics[width=0.95\linewidth]{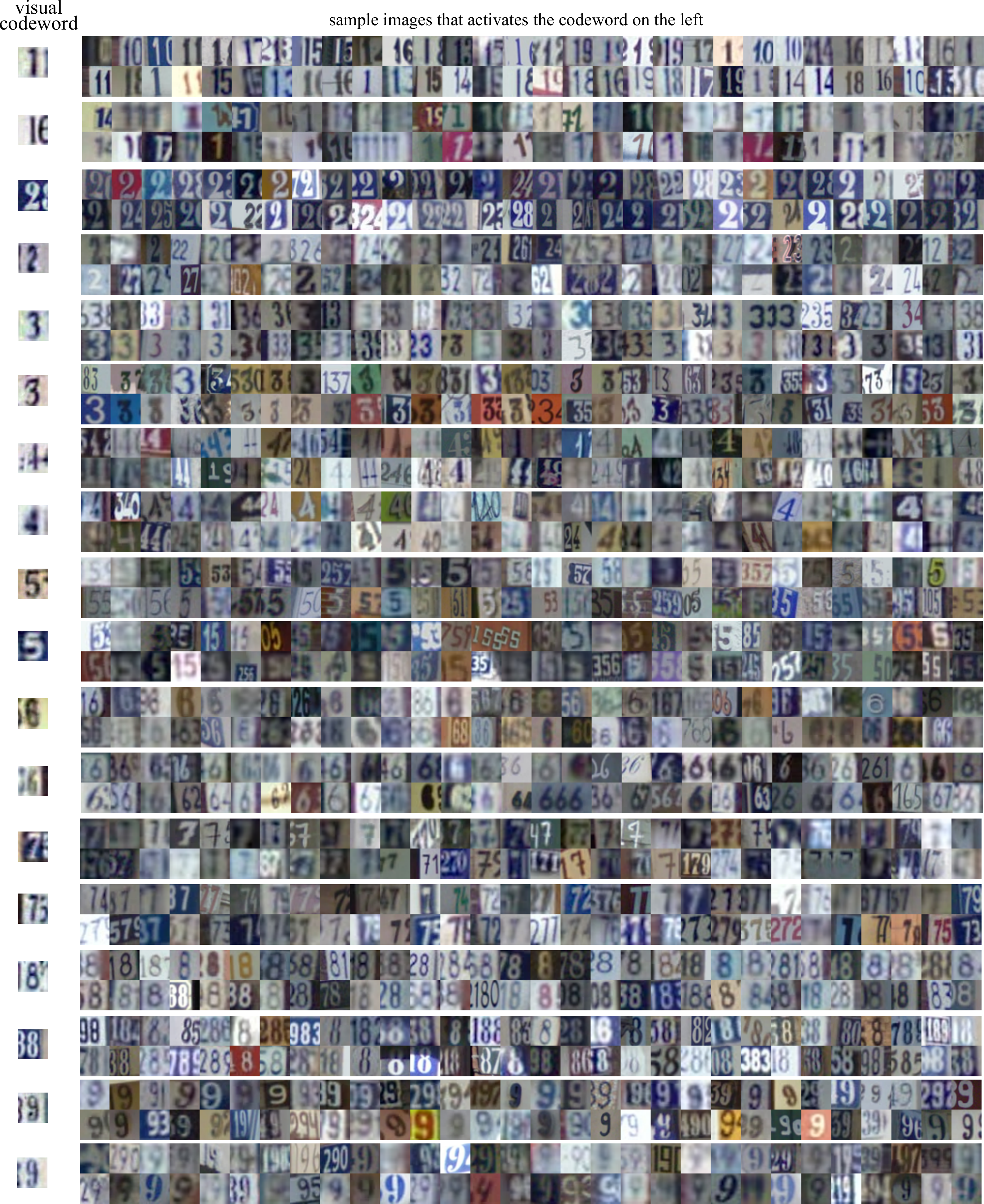} 
	
	\caption{{\bf  Dataset images from SVHN that activate a particular codeword.} Each block has two columns. The left one shows the visual codeword and the right one  the dataset images that activated the codeword on the left.  Note that there is a clear semantic similarity between the images and the codeword, in terms of class label, but also with respect to style and orientation.}
	\vspace{+0.05in}\label{fig:assgn_svhn}
	
\end{figure*}

\begin{figure*} [h] \centering
	
	\includegraphics[width=0.95\linewidth]{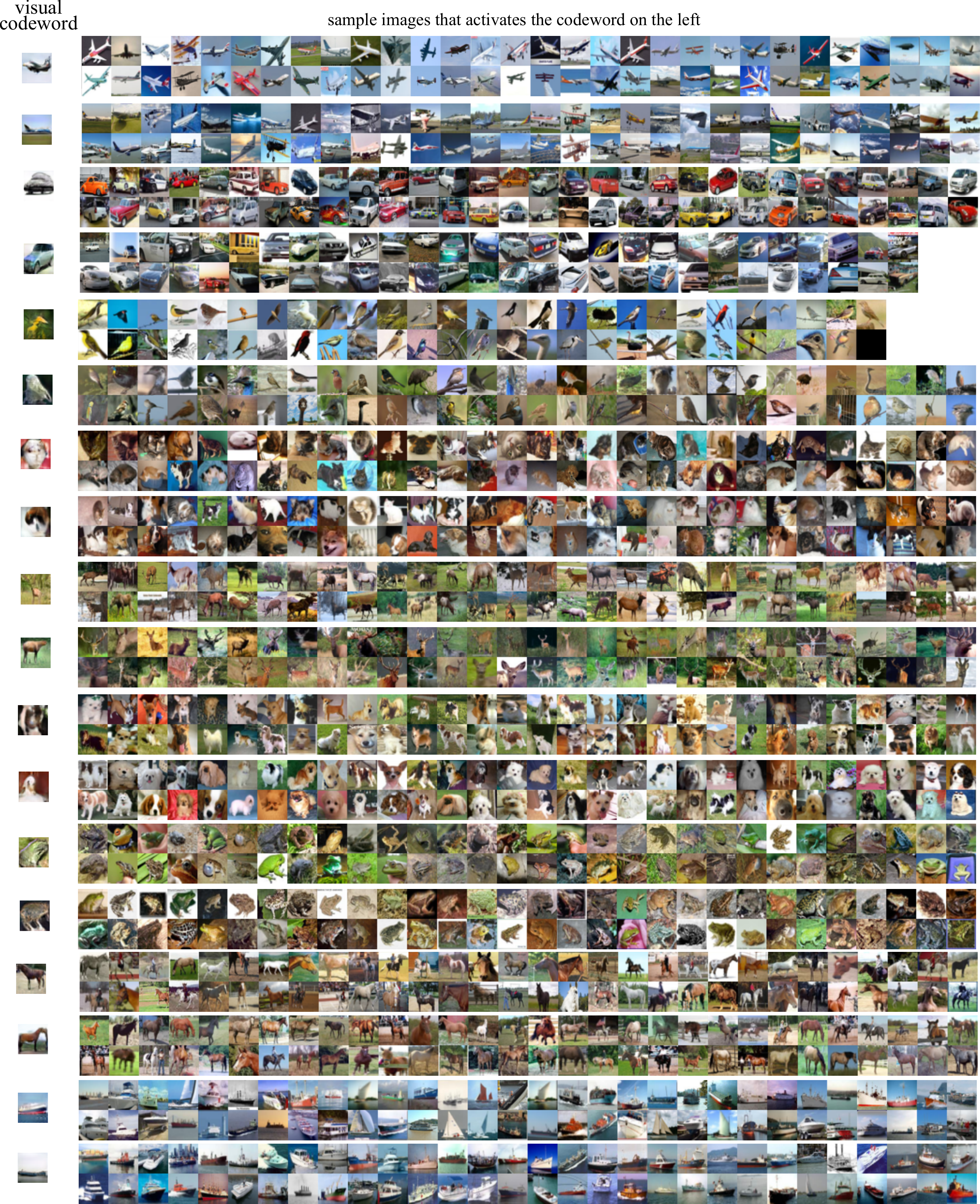} 
	
	\caption{{\bf  Dataset images from CIFAR-10 that activate a particular codeword.} Each block has two columns. The left one shows the visual codeword and the right one  the dataset images that activated the codeword on the left.  Note that there is a clear semantic similarity between the images and the codeword in terms of class label.}
	\vspace{+0.05in}\label{fig:assgn_cifar}
	
\end{figure*}

\begin{figure*} [h] \centering
	\includegraphics[width=0.95\linewidth]{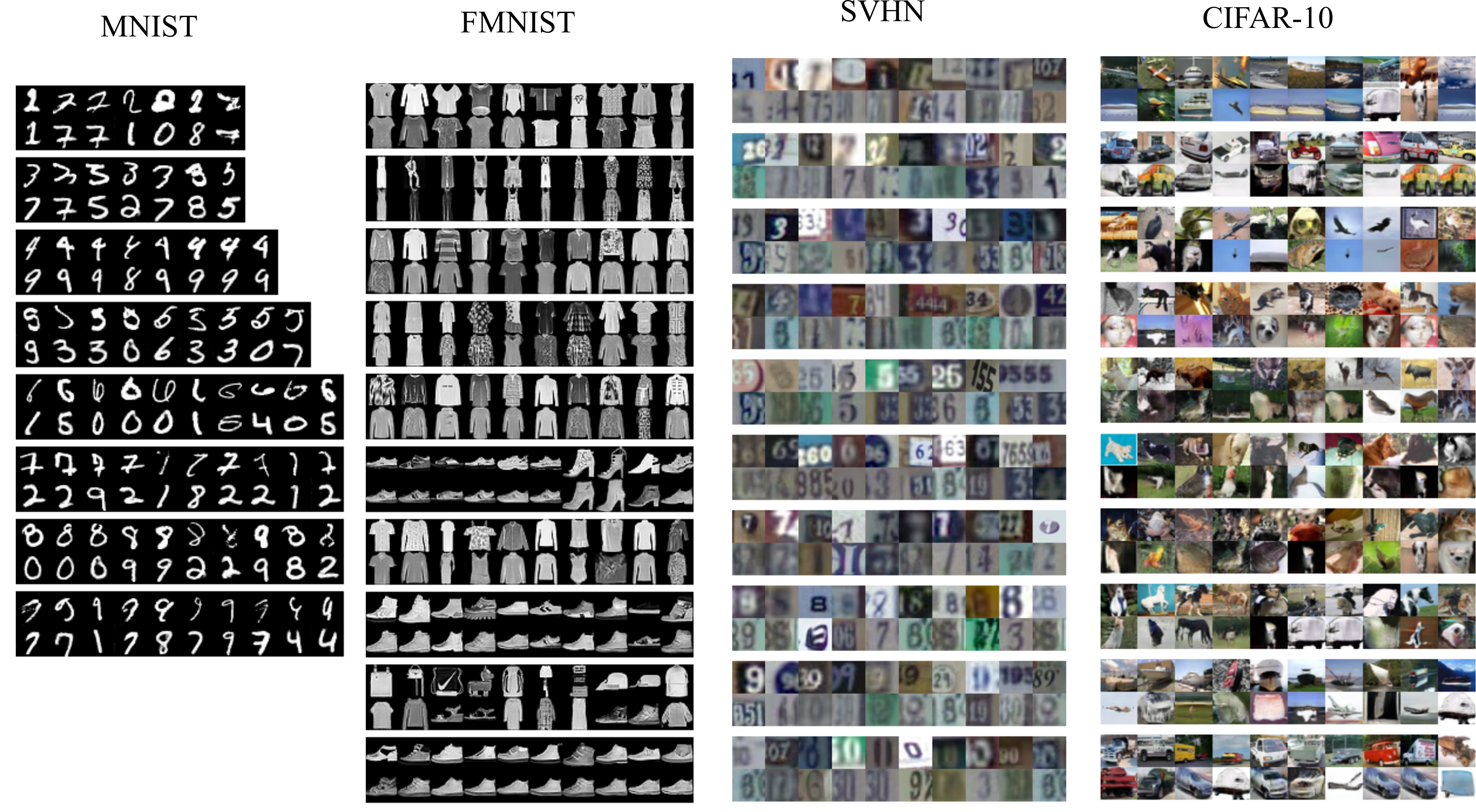} 
	\caption{{\bf Most highly activated codewords of incorrectly classified  test samples on various datasets.} Each two rows within a block show the input image (top) of a specific class and the corresponding codeword (bottom). Note that the images and their codeword are dissimilar in most cases.}\label{fig:codewords_incorrect}
\end{figure*}

We visualize the learned codewords on various datasets in Fig.~\ref{fig:codewords_full}. We also provide additional visualizations of  the activated visual codewords for normal  samples of MNIST, FMNIST, SVHN and CIFAR-10 in Fig.~\ref{fig:normal_all}. We further visualize the images assigned to a particular codeword in Figs.~\ref{fig:assgn_mnist}, ~\ref{fig:assgn_fmnist}, ~\ref{fig:assgn_svhn}  and ~\ref{fig:assgn_cifar} for MNIST, FMNIST, SVHN and CIFAR-10 respectively. Finally, the misclassified normal samples for all datasets are shown in Fig.~\ref{fig:codewords_incorrect}.

\section{Adversarial Attack Implementations}

We used the Cleverhans library\footnote{https://github.com/tensorflow/cleverhans} to generate adversarial images following the same settings as in~\cite{ma2018characterizing}. In Table~\ref{table:L2values}, we report the mean $L_2$ perturbation of the adversarial images for different attack methods. Note that FGSM and BIM-b tend to yield larger perturbations than BIM-a and CW.

\begin{table}[h!]
	
	\resizebox{0.9\linewidth}{!}{%
		
		\begin{tabular}{|c|c|cccc|}
			\hline
			\multirow{1}{*}{Dataset}  & Model & FGSM & BIM-a & BIM-b & CW  \\ \hline
			
			\multirow{2}{*}{MNIST} 
			& Base    &7.45&2.67&6.73&1.55\\
			&   BoW  &8.01&3.30&3.45&0.04\\ \hline
			
			\multirow{2}{*}{F-MNIST} 
			& Base     &4.61&1.01&4.07&0.90\\
			&   BoW  &4.62&1.78&2.07&0.97\\ \hline
			
			\multirow{2}{*}{CIFAR-10} 
			& Base    & 2.74&0.46&2.12&0.35\\
			
			&   BoW  &2.73&0.66&1.54&0.45\\ \hline
			
			\multirow{2}{*}{SVHN} 
			& Base     &7.08&1.02&6.16&1.16\\
			&   BoW  &7.09&1.75&3.48&1.52\\ \hline
			
		\end{tabular}%
	}
	\caption{\textbf{Mean $L_2$ perturbation} of the adversarial images on the base and BoW networks with different attack strategies.}  
	\label{table:L2values}
	
\end{table}


\section{Influence of the Number of Codewords }

We evaluate the influence of the hyper-parameter $K$ in Eq. 2 of the main paper, which defines the number of codewords in our BoW models.To this end, in Table~\ref{table:param_K_results}, we report the detection AUROC for different values of $K$ with FGSM 
attack on various datasets with evaluation Strategy 2. Perhaps surprisingly, with as few as $K=50$ clusters (5 clusters per class), we still obtain high AUROCs. 


\begin{table}[h!]
	
	\centering
	\resizebox{0.6\linewidth}{!}{%
		
		\begin{tabular}{|c|c|c|c|c|}
			\hline
			\multirow{1}{*}{$K$}  &  50 & 100 & 500 & 1000  \\ \hline
			
			MNIST    & 99.95 &100.0&100.0&100.0\\ \hline
			FMNIST   &100.0&100.0&100.0&100.0\\ \hline
			SVHN       & 99.33  & 100.0& 100.0& 100.0\\ \hline

		\end{tabular}%
	}
	\caption{\textbf{Influence of $K$} on detection AUROC}
	\label{table:param_K_results}

\end{table}


\section{Additional Results on Adversarial Samples}

\begin{figure*} [h] \centering
	\vspace{-0.1in}
	\subfigure[FGSM]
	{
		\includegraphics[width=0.45\linewidth]{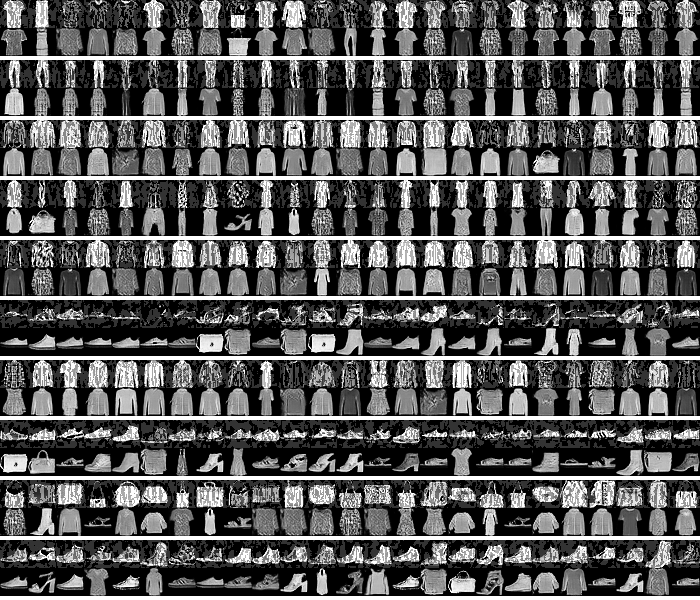} } 
	\,
	\subfigure[BIM-a]
	{
		\includegraphics[width=0.45\linewidth]{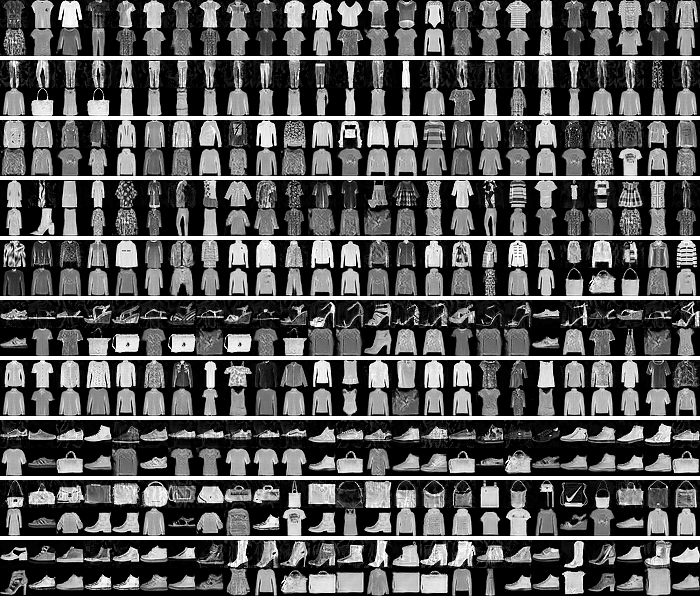} }
	\vspace{0.1cm}
	\subfigure[BIM-b]
	{
		\includegraphics[width=0.45\linewidth]{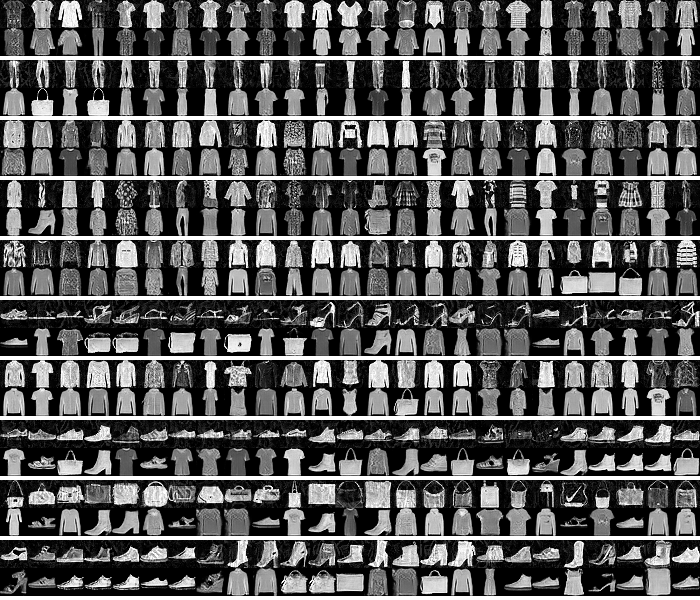}}
	\,
	\subfigure[CW]
	{
		\includegraphics[width=0.45\linewidth]{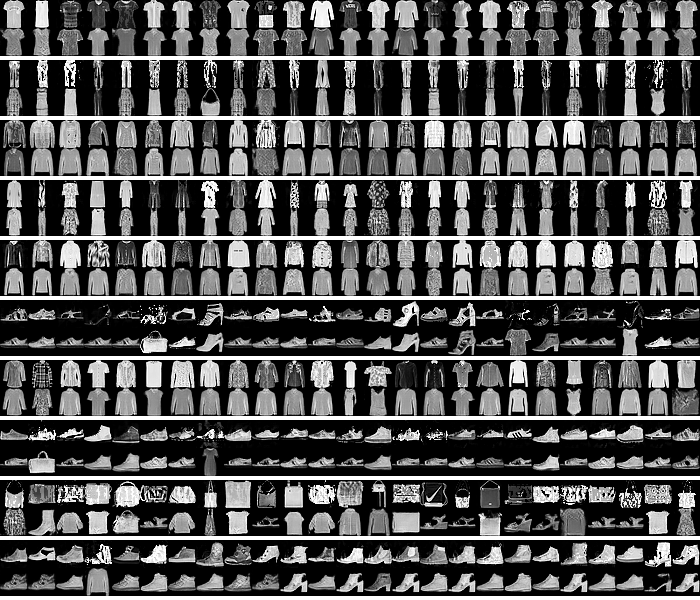}
	}
	
	\caption{{\bf Most highly activated codewords of adversarial samples obtained with different attack strategies on FMNIST test data.} Each two rows within a block show the input image (top) and corresponding codeword (bottom). Note that the adversarial images and their codeword are dissimilar in most cases.}
	\vspace{+0.05in}
	\label{fig:fmnist_adv_all}
\end{figure*}

\begin{figure*} [h] \centering
	\vspace{-0.1in}
	\subfigure[FGSM]
	{
		\includegraphics[width=0.45\linewidth]{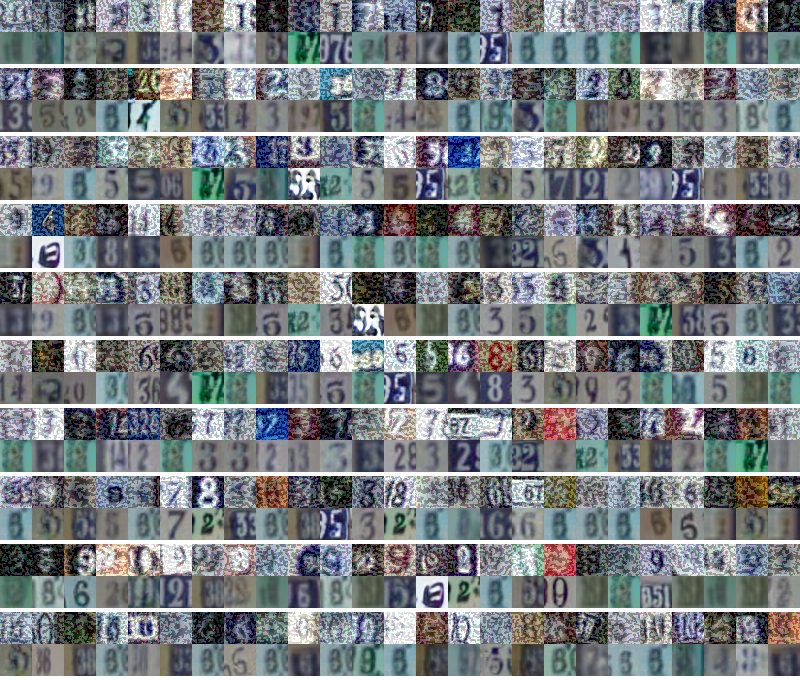} } 
	\,
	\subfigure[BIM-a]
	{
		\includegraphics[width=0.45\linewidth]{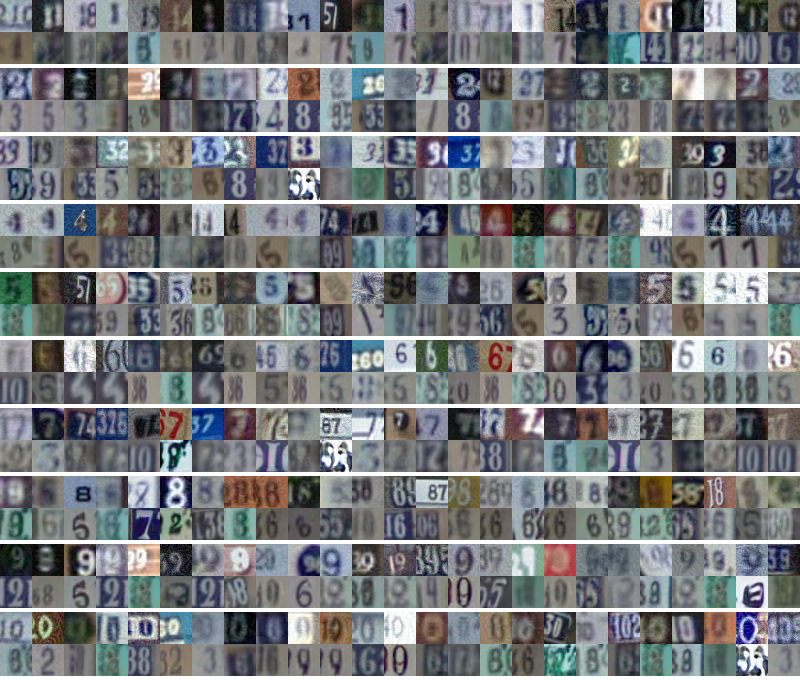} }
	\vspace{0.1cm}
	\subfigure[BIM-b]
	{
		\includegraphics[width=0.45\linewidth]{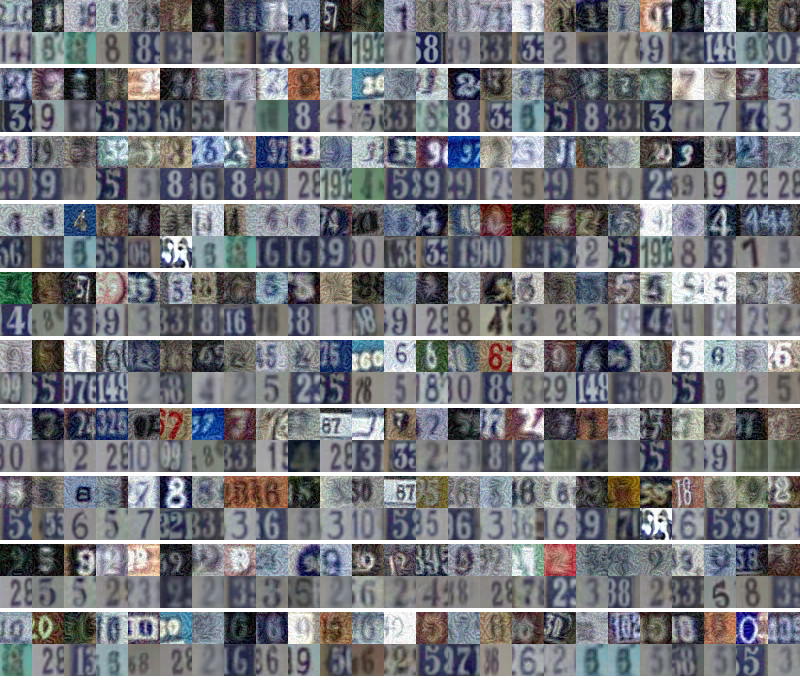}}
	\,
	\subfigure[CW]
	{
		\includegraphics[width=0.45\linewidth]{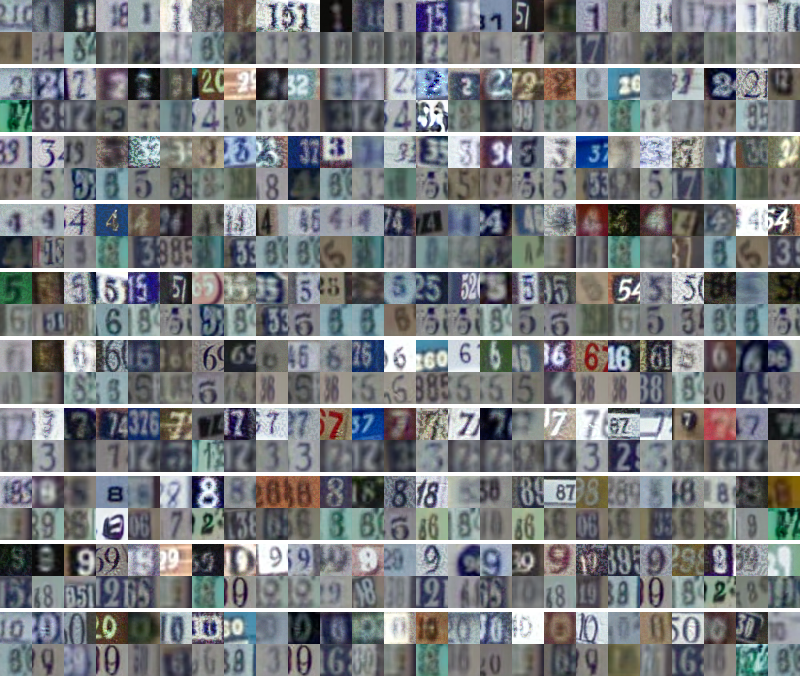}
	}
	
	\caption{{\bf Most highly activated codewords of adversarial samples obtained with different attack strategies on SVHN test data.} Each two rows within a block show the input image (top) and corresponding codeword (bottom). Note that the adversarial images and their codeword are dissimilar in most cases.}
	\vspace{+0.05in}
	\label{fig:svhn_adv_all}
\end{figure*}

\begin{figure*} [h] \centering
	\vspace{-0.1in}
	\subfigure[FGSM]
	{
		\includegraphics[width=0.45\linewidth]{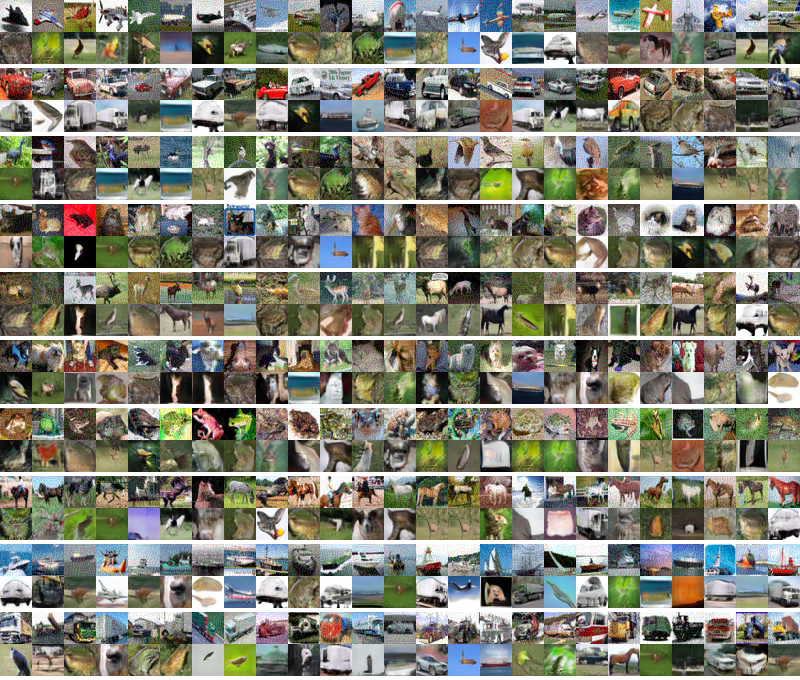} } 
	\,
	\subfigure[BIM-a]
	{
		\includegraphics[width=0.45\linewidth]{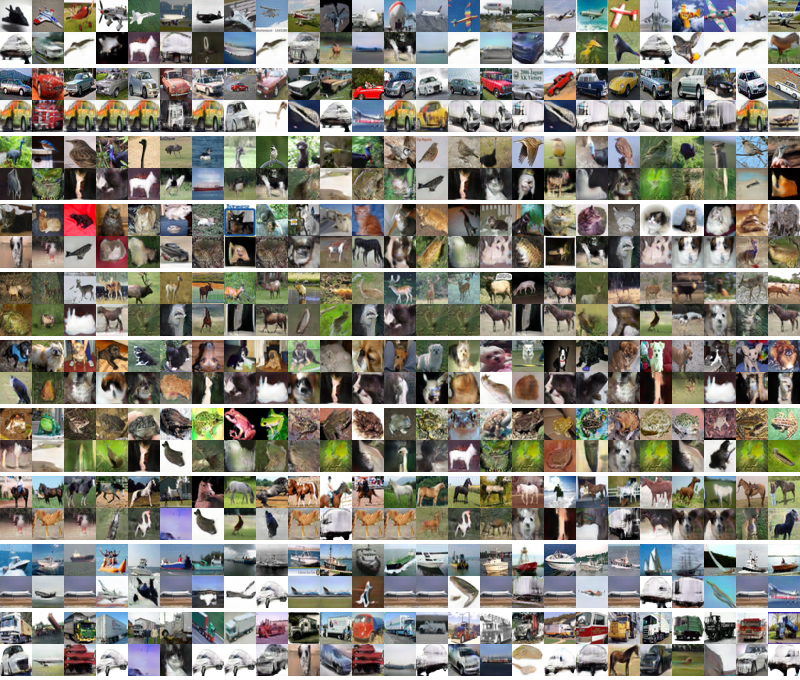} }
	\vspace{0.1cm}
	\subfigure[BIM-b]
	{
		\includegraphics[width=0.45\linewidth]{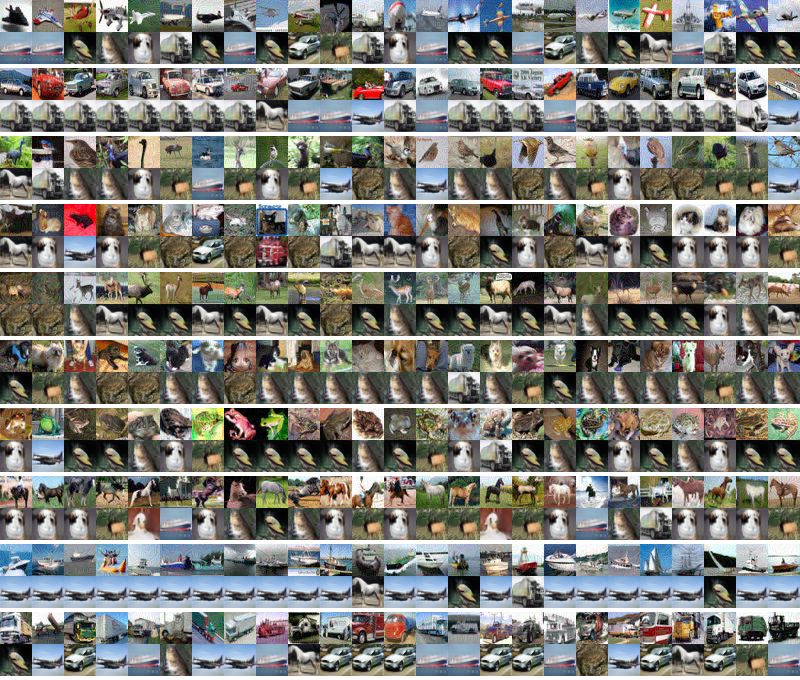}}
	\,
	\subfigure[CW]
	{
		\includegraphics[width=0.45\linewidth]{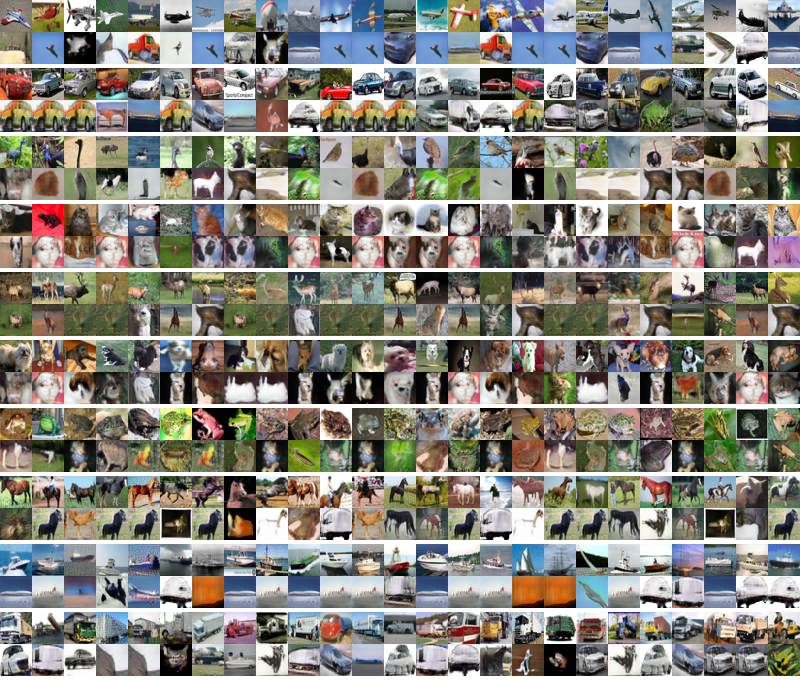}
	}
	
	\caption{{\bf Most highly activated codewords of adversarial samples obtained with different attack strategies on CIFAR-10 test data.} Each two rows within a block show the input image (top) and corresponding codeword (bottom). Note that the adversarial images and their codeword are dissimilar in most cases.}
	\vspace{+0.05in}
	\label{fig:cifar_adv_all}
\end{figure*}

We also provide additional visualizations of  activated visual codewords for adversarial samples of MNIST, FMNIST, SVHN and CIFAR-10 in Figs.~\ref{fig:mnist_adv_all},~\ref{fig:fmnist_adv_all},~\ref{fig:svhn_adv_all} and~\ref{fig:cifar_adv_all}, respectively.


\clearpage


\clearpage


\clearpage

\clearpage

\clearpage

\clearpage


\clearpage




\begin{figure*} [h] \centering
	\vspace{-0.1in}
	\subfigure[FGSM]
	{
		\includegraphics[width=0.45\linewidth]{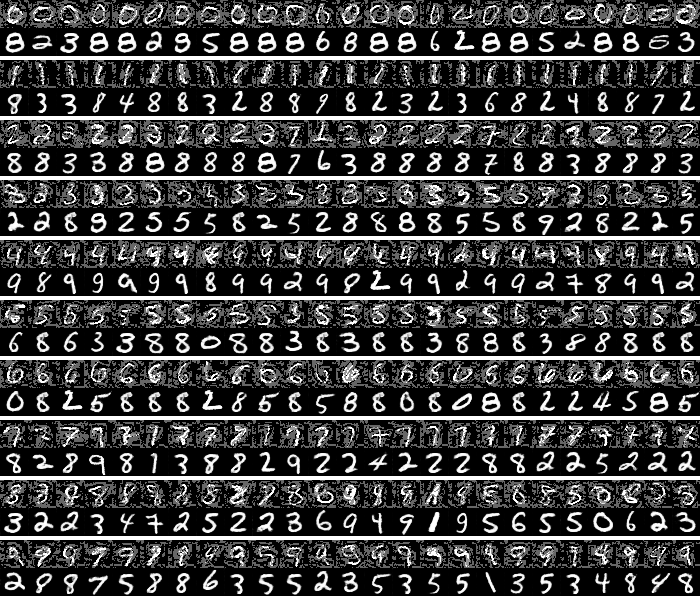} } 
	\,
	\subfigure[BIM-a]
	{
		\includegraphics[width=0.45\linewidth]{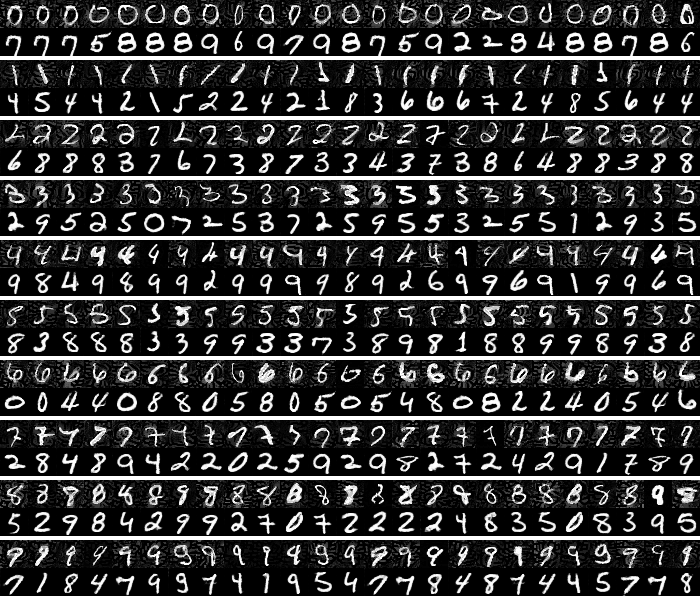} }
	\vspace{0.1cm}
	\subfigure[BIM-b]
	{
		\includegraphics[width=0.45\linewidth]{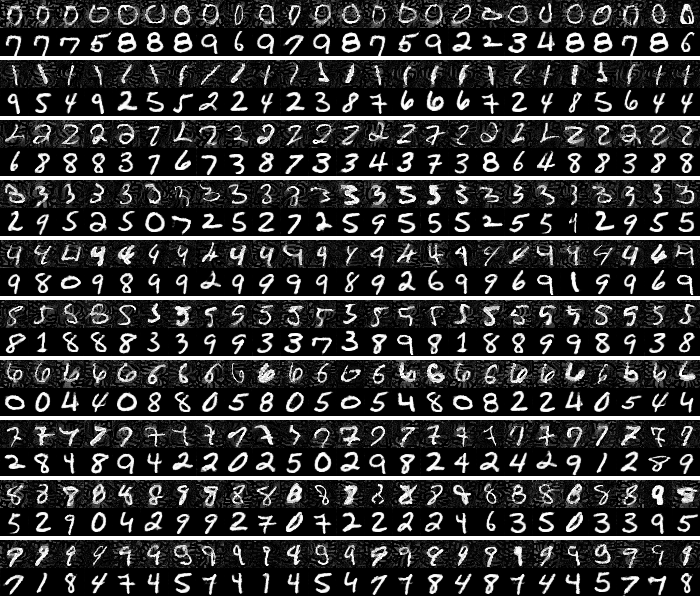}}
	\,
	\subfigure[CW]
	{
		\includegraphics[width=0.45\linewidth]{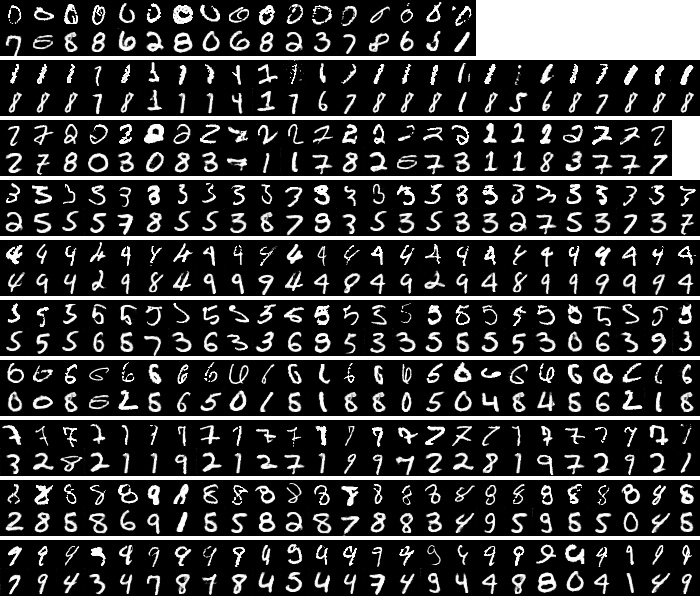}
	}
	
	\caption{{\bf Most highly activated codewords of adversarial samples obtained with different attack strategies on MNIST test data.} Each two rows within a block show the input image (top) and corresponding codeword (bottom). Note that the adversarial images and their codeword are dissimilar in most cases. Note that, in the CW case, we have fewer samples in some of the classes because of the low success rate of this attack.}
	\vspace{+0.05in}
	\label{fig:mnist_adv_all}
\end{figure*}

\comment{

\begin{figure*} [h] \centering
	\vspace{-0.1in}
	\subfigure[FGSM]
	{
		\includegraphics[width=0.45\linewidth]{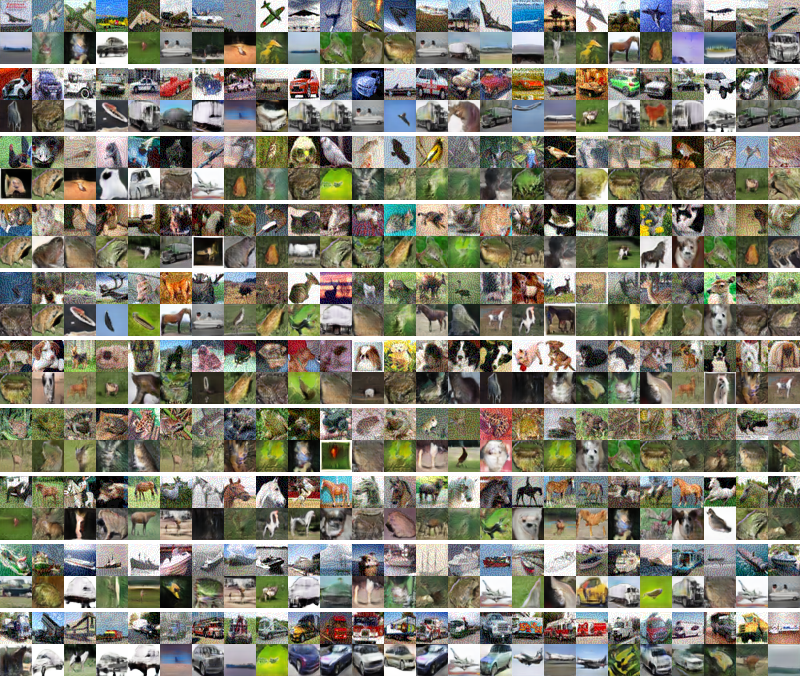} } 
	\,
	\subfigure[BIM-a]
	{
		\includegraphics[width=0.45\linewidth]{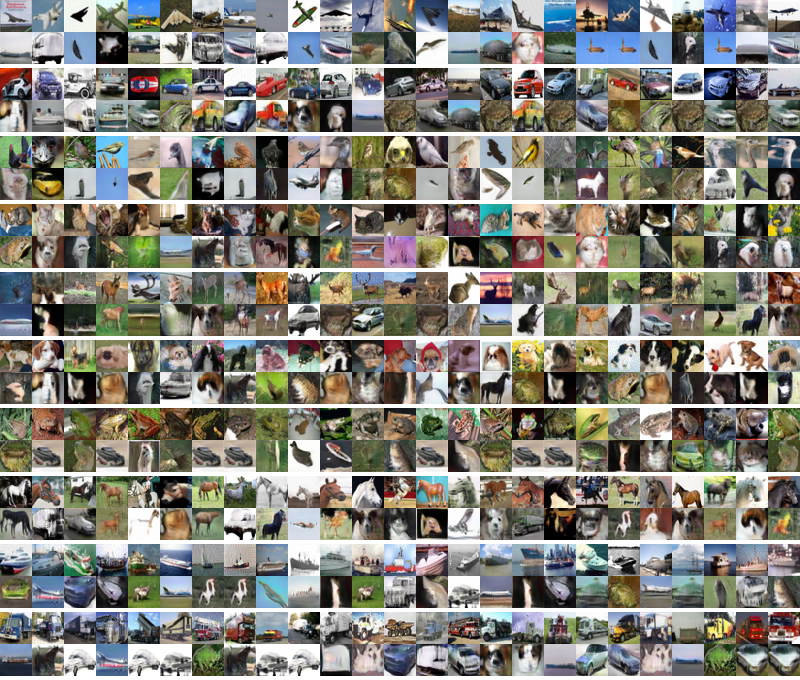} }
	\vspace{0.1cm}
	\subfigure[BIM-b]
	{
		\includegraphics[width=0.45\linewidth]{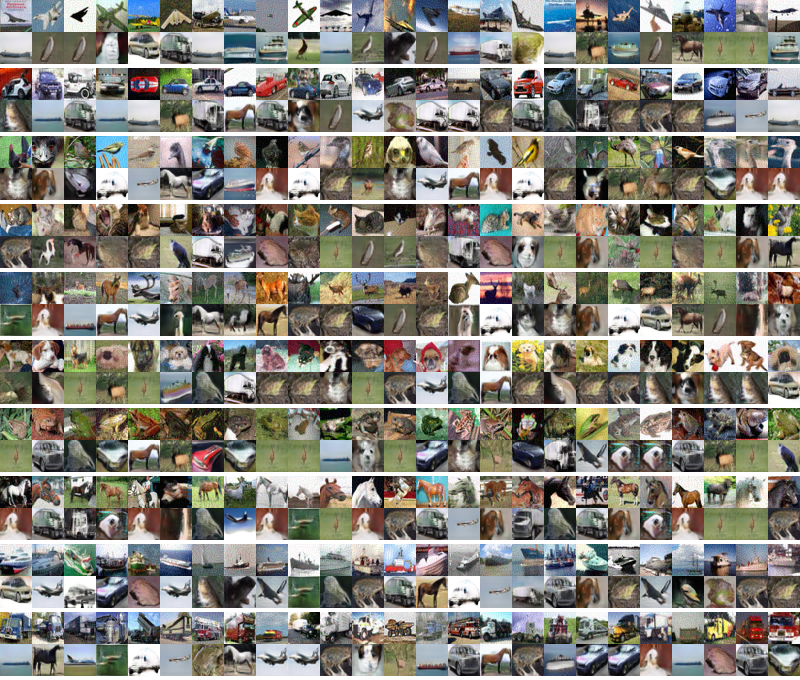}}
	\,
	\subfigure[CW]
	{
		\includegraphics[width=0.45\linewidth]{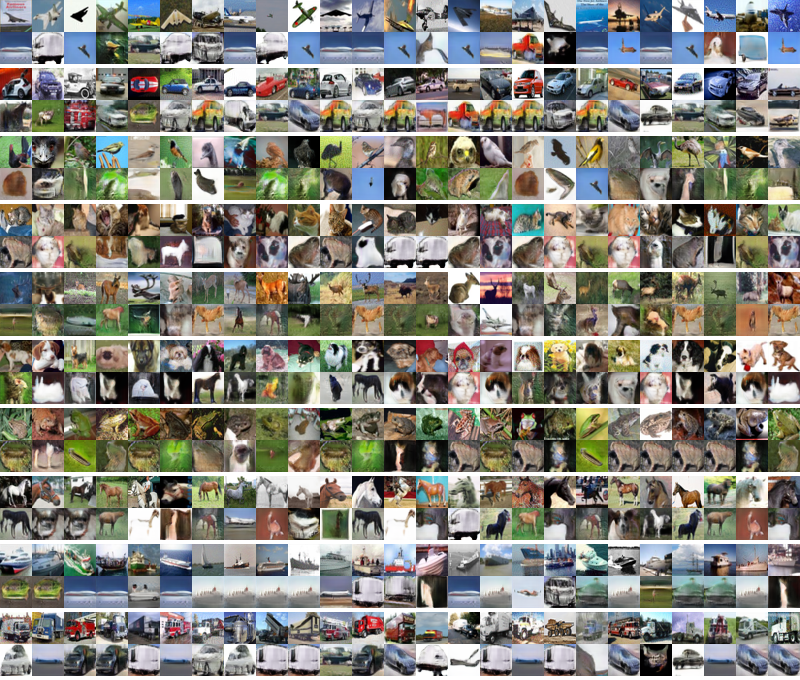}
	}
	
	\caption{{\bf Most highly activated codewords of adversarial samples obtained with different attack strategies on CIFAR10 test data.} Each two rows within a block show the input image (top) and corresponding codeword (bottom). Note that the adversarial images and their codeword are dissimilar in most cases.}
	\vspace{+0.05in}
	\label{fig:mnist_adv_all}
\end{figure*}

}

\end{document}